\documentclass{article}

\usepackage{microtype}
\usepackage{graphicx}
\usepackage{subfigure}
\usepackage{subcaption}
\usepackage{booktabs} 

\usepackage{hyperref}

\usepackage[accepted]{icml2025}

\usepackage{amsmath}
\usepackage{amssymb}
\usepackage{mathtools}
\usepackage{amsthm}
\usepackage{xspace}
\usepackage{bbm}
\usepackage{multirow}
\usepackage{bm}
\usepackage[usestackEOL]{stackengine}
\usepackage{enumitem}
\usepackage{svg}

\usepackage[capitalize,noabbrev]{cleveref}

\usepackage{pifont}
\newcommand{\cmark}{\ding{51}}%
\newcommand{\xmark}{\ding{55}}%

\theoremstyle{plain}

\theoremstyle{definition}

\theoremstyle{remark}

\usepackage[textsize=tisny]{todonotes}

\newcommand{\ours}{DiffMS\xspace}

\icmltitlerunning{DiffMS: Diffusion Generation of Molecules Conditioned on Mass Spectra}

\begin{document}

\twocolumn[
\icmltitle{DiffMS: Diffusion Generation of Molecules Conditioned on Mass Spectra}




\begin{icmlauthorlist}
\icmlauthor{Montgomery Bohde}{yyy,xxx}
\icmlauthor{Mrunali Manjrekar}{yyy}
\icmlauthor{Runzhong Wang}{yyy}
\icmlauthor{Shuiwang Ji}{xxx}
\icmlauthor{Connor W. Coley}{yyy}
\end{icmlauthorlist}

\icmlaffiliation{yyy}{Massachusetts Institute of Technology, Cambridge, MA, United States}
\icmlaffiliation{xxx}{Texas A\&M University, College Station, TX, United States}

\icmlcorrespondingauthor{Connor W. Coley}{ccoley@mit.edu}


\vskip 0.3in
]



\printAffiliationsAndNotice{} 

\begin{abstract}
Mass spectrometry plays a fundamental role in elucidating the structures of unknown molecules and subsequent scientific discoveries. One formulation of the structure elucidation task is the conditional \emph{de novo} generation of molecular structure given a mass spectrum. Toward a more accurate and efficient scientific discovery pipeline for small molecules, we present DiffMS, a formula-restricted encoder-decoder generative network that achieves state-of-the-art performance on this task. The encoder utilizes a transformer architecture and models mass spectra domain knowledge such as peak formulae and neutral losses, and the decoder is a discrete graph diffusion model restricted by the heavy-atom composition of a known chemical formula. To develop a robust decoder that bridges latent embeddings and molecular structures, we pretrain the diffusion decoder with fingerprint-structure pairs, which are available in virtually infinite quantities, compared to structure-spectrum pairs that number in the tens of thousands. 
Extensive experiments on established benchmarks show that DiffMS outperforms existing models on \emph{de novo} molecule generation. We provide several ablations to demonstrate the effectiveness of our diffusion and pretraining approaches and show consistent performance scaling with increasing pretraining dataset size. DiffMS code is publicly available at \url{https://github.com/coleygroup/DiffMS}.
\end{abstract}

\begin{figure}
    \centering
    \includegraphics[width=\linewidth]{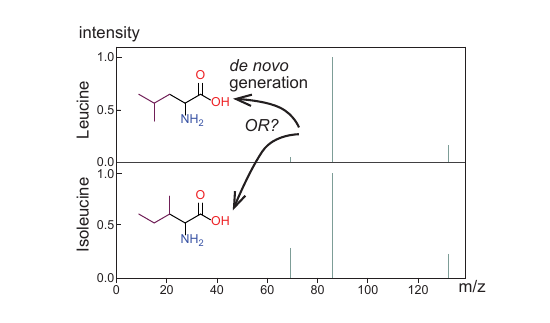}
    \vspace{-20pt}
    \caption{\textit{De novo} structure generation from LC-MS/MS faces ambiguity when isobaric or isomeric compounds yield similar fragmentation spectra. In this case, the experimental spectra for leucine and isoleucine from \citet{nist_database} are essentially indistinguishable. It is one of many examples demonstrating that the identification of the exact structure is desirable but challenging. 
    }
    \label{fig:leu-ile}
\end{figure}

\begin{figure*}[tb!]
    \centering
    \includegraphics[width=\linewidth]{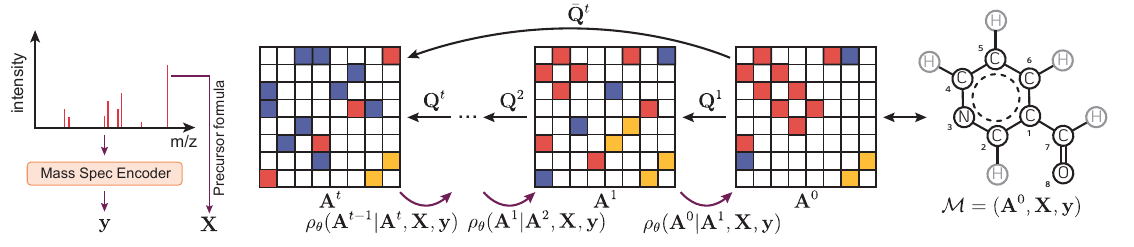}
    \caption{DiffMS tackles \textit{de novo} molecular generation from mass spectra. We embed mass spectrum features with a transformer encoder, and assume the chemical formula is determined by off-the-shelf tools~\citep{goldman2023mist-cf,bocker2016fragmentation} so that the numbers and types of heavy atoms (i.e.\ nodes in the molecular graph) is constrained. The molecular structure is represented as an adjacency matrix with one-hot encoded bond types, which in this example are single (blue), double (yellow), aromatic bonds (red) and no bond (white). The target molecular structure is generated starting from a randomly initialized adjacency matrix, which is denoised through a discrete diffusion process~\citep{vignac2023digress}. The trajectory used for training is created by randomly disturbing the true structure $t$ times.}
    \label{fig:task_overview}
\end{figure*}

\section{Introduction}
\label{intro}

Mass spectrometry (MS) is a fundamental part of the analytical chemistry toolkit that can assist in the identification of unknown compounds of interest collected from experiments. Tandem mass spectrometry (MS/MS) in combination with liquid chromatography (LC) enables information-rich, high-throughput profiling of compounds, wherein complex experimental mixtures are separated in two dimensions, first by retention time (from chromatography) and then by molecule m/z (mass-to-charge ratio) in the first MS (MS1) stage. 
Each ``precursor'' molecule is then individually passed through the second MS stage (MS2) where it undergoes collision-induced dissociation and is split into a set of charged molecular fragments, each with a corresponding m/z and an intensity. 
Modern LC-MS/MS has enabled the discovery of many new compounds of interest, such as identifying novel bile acids in microbiome study~\citep{quinn2020global}, uncovering a tire rubber-derived chemical that is toxic to coho salmon~\citep{tian2021ubiquitous}. There is also a growing interest in increased throughput with MS technologies, such as a high-throughput analysis of chemical reactions~\citep{hu2024continuous} and a systematic discovery pipeline for human metabolites~\citep{gentry2024reverse}.


From MS1 and MS2 data alone, elucidation of the chemical structure(s) present in the original experimental sample remains 
challenging. Many yet-to-be-discovered metabolites have structures that do not exist in standard virtual chemical libraries (PubChem, Human Metabolome Database, etc.). The majority of observed spectra in MS-based metabolomics campaigns remain unidentified and are characterized as ``metabolite dark matter''~\citep{bittremieux2022critical}. The difficulty of the elucidation task comes from both computation and chemistry. In terms of computation, there is a large set of possible substructures to explain each measurement, creating an exponential number of structure candidates for the overall mass spectrum, i.e., an NP-hard combinatorial optimization. From the chemistry perspective, a standalone mass spectrum may be insufficient to determine a unique structure because of the ionization and fragmentation mechanisms of the instrument; as shown in Fig.~\ref{fig:leu-ile}, the spectra of two isomeric amino acids are nearly identical. While we would like to be able to determine the exact structure, from an application perspective, generating similar but not exactly matching structures is still useful to domain experts to narrow down the chemical space. 


Machine learning methods have recently taken root in this space to address two key challenges in particular: 1) to learn how to fragment a given molecule, and predict the resultant mass spectrum, known as ``forward'' MS simulation ~\citep{murphy2023efficiently,goldman2023scarf,goldman2024iceberg,young2024fragnnet,young2024massformer,nowatzky2024fiora} and 2) to take an experimental spectrum and predict the corresponding structure or a description thereof, typically as a fingerprint, SMILES, or graph representation, known as ``inverse'' methods~\citep{duhrkop2015csifingerid,stravs2022msnovelist, butler2023ms2mol,litsa2023spec2mol,goldman2023mist}. 



In this paper, we focus on the ``inverse'' MS problem and develop a novel machine learning framework for chemical structure generation given a mass spectrum, sometimes also described as \emph{de novo} generation. 
A recent study of \emph{de novo} generation from MS shows that all of the methods tested suffer from near-zero structural accuracy~\citep{bushuiev2024massspecgymbenchmarkdiscoveryidentification}. 
Among prior approaches to this task are language models that are trained to convert tokenized m/z values and intensities as inputs to SMILES strings as outputs~\citep{butler2023ms2mol,litsa2023spec2mol}; however, these autoregressive language models do not capture the permutation-invariant nature of mass spectra and molecules, nor do they incorporate chemical formula constraints as helpful prior knowledge. 
Another family of approaches utilizes intermediate representations such as scaffolds~\citep{wang2025madgenmassspecattendsnovo} or fingerprints~\citep{stravs2022msnovelist} before generating chemical structures, which are arguably more chemically interpretable and leverage additional amounts of structure-only data available, 
but have not necessarily led to significant performance improvement on benchmarks.
Compared to the complete structural elucidation challenge of the ``inverse'' MS problem, 
the chemical formula of the unknown molecule is usually easier to determine by off-the-shelf tools from MS1 and MS2 data, utilizing tools such as SIRIUS \cite{bocker2016fragmentation}, BUDDY~\cite{Xing2023-buddy}, or MIST-CF \cite{goldman2023mist-cf}. 
One insight of our work is to use those available tools by taking the chemical formula as given and developing a formula-constrained (i.e., heavy atom-constrained) generation pipeline. 
We also find it beneficial to exploit intermediate chemical representations to enable a scaled-up pretraining stage with theoretically unlimited fingerprint-structure pairs. 



To this end, we present \ours, a permutation-invariant diffusion model trained end-to-end for molecule generation conditioned on mass spectra (Fig.~\ref{fig:task_overview}).
\ours has an encoder-decoder architecture that builds upon modern transformers~\citep{vaswani2017attention} and discrete graph diffusion~\citep{vignac2023digress}. 
For the encoder, we adopt the formula transformer from MIST~\citep{goldman2023mist} with pairwise modeling of neutral losses as a domain-informed inductive bias. 
For the decoder, we build upon the DiGress graph diffusion model~\citep{vignac2023digress} using chemical formula constraints and embeddings extracted from the formula transformer as the condition to generate target molecules. 
We provide empirical validation of our end-to-end model on established mass spectra \emph{de novo} generation benchmarks~\citep{duhrkop2021canopus, bushuiev2024massspecgymbenchmarkdiscoveryidentification}. Additional ablation studies demonstrate the effectiveness of our pretraining-finetuning framework. 

Our contributions are summarized as follows:
\begin{enumerate}
    \item We present \ours, the first conditional molecular generator with formula constraints for structural elucidation from mass spectra. We demonstrate discrete diffusion as a natural methodology for conditional molecular generation that natively handles predefined heavy-atom composition and accounts for the underspecification of conditioning (i.e., the one-to-many mapping from spectrum to structure illustrated by Fig.~\ref{fig:leu-ile}). 
    
    \item We propose a pretraining-finetuning framework for training \ours that makes use of virtual chemical libraries with self-labeled structural conditions. 
    Specifically, the diffusion decoder is trained on a large-scale dataset with 2.8M fingerprint-structure pairs. Our ablation studies show that downstream performance scales well with increasing fingerprint-structure pretraining dataset size, providing a promising avenue to scale the performance. 
    We also pretrain the spectrum encoder to predict fingerprints from spectra embeddings to further boost performance of the end-to-end finetuned model. 
    
    
    \item On established benchmarks for \emph{de novo} structure elucidation, \ours demonstrates superior performance compared to all existing baselines, achieving improved annotation accuracy and better structural similarity to the true molecule. While \emph{de novo} generation of the exact molecular structure remains challenging, structurally close matches can offer valuable insights for domain experts~\citep{butler2023ms2mol}. The broad applicability of MS underscores the potential impact of \ours in advancing research in chemical and biological discovery. 
    
    

\end{enumerate}

\section{Background and Related Work}

\subsection{Conditional generative molecular design}

Unconditional molecular generation has been well-explored in the context of AI for chemistry~\citep{zhang2023artificial}, with methods such as \citet{gomez2018automatic,segler2018generating} leveraging autoregressive sampling with language decoders to generate SMILES representations of molecules as well as GNN architectures that generate molecular graphs atom-by-atom for either 2-dimensional~\citep{liu2018constrained,li2018learning,simonovsky2018graphvae} or 3-dimensional~\citep{flam2022scalable,adams2022equivariant,luo2022autoregressive,liu2022generating3dmoleculestarget} graphs. Recently, \citet{vignac2023digress} developed DiGress, a non-autoregressive generative model based on discrete graph diffusion. The target spaces of these generative models are generally unconditioned or loosely conditioned, for example, to generate drug-like molecules~\citep{luo20213d} or molecules with certain conformations~\citep{roney2022generating}.

In the context of \textit{de novo} structural generation, however, molecular generation must be strongly conditioned on the spectral information, i.e., the fragmentation pattern itself and an inferred chemical formula. There are some efforts that try to generate structures from molecular fingerprints, which is another form of strong structural condition, including Neuraldecipher~\citep{le2020neuraldecipher} that learns how to decode SMILES strings from molecular fingerprints, as well as MSNovelist~\citep{stravs2022msnovelist}, which proposes a fingerprint-to-SMILES long short-term memory (LSTM) network. 

Both of these methods use autoregressive models that cannot strictly enforce formula constraints, while in MS, the chemical formula of the target molecule is an important inductive bias that limits the target space. In this paper, we identify discrete graph diffusion as a natural choice to incorporate formula constraints and improve \citet{vignac2023digress}, expanding the suite of methods in conditional molecular generation.

\subsection{Inverse models for structure elucidation from spectra}

Inverse models take an experimental spectrum as input and predict a relevant representation of the structure: typically, the molecular graph itself, a SMILES string, or a molecular fingerprint. 
DENDRAL, arguably the first expert system that applied AI to science, focuses on structural elucidation from mass spectrometry data~\citep{lindsay1980dendral}. 
Recent years have seen the adoption of machine learning for a new class of inverse MS models, such as for spectrum-to-fingerprint predictions, involving either support vector machines, as in CSI:FingerID~\citep{duhrkop2015csifingerid}; or deep learning with transformers, as in MIST~\citep{goldman2023mist}. The fingerprint, which is a binary encoding of the structure, can be further used to rank candidate structures from a chemical library. Similar elucidation goals have been pursued with other types of analytical spectra such as nuclear magnetic resonance (NMR)~\citep{alberts2023learning}. 

However, the elucidation of structures that do not necessarily exist in any virtual chemical library requires generative techniques rather than retrieval-based techniques. MSNovelist~\citep{stravs2022msnovelist} builds an autoregressive fingerprint-to-molecule model that takes fingerprint predictions returned by CSI-FingerID and generates SMILES strings, with a decoding process that utilizes the molecular formula of the candidate compound as inferred from tools such as SIRIUS\cite{bocker2016fragmentation} or MIST-CF \cite{goldman2023mist-cf}. 
Spec2Mol~\citep{litsa2023spec2mol} develops a SMILES autoencoder and trains a spectrum CNN encoder model, with up to four spectral channels to accept spectra collected in low or high energy and positive or negative mode, that tries to predict the corresponding SMILES embedding from the spectrum. 
MassGenie~\citep{shrivastava2021massgenie} is an orthogonal effort that uses forward MS models~\citep{allen2015cfm-id,goldman2024iceberg} to augment training datasets with \emph{in silico} reference spectra.
Toward an end-to-end pipeline for molecular generation from mass spectra, which is the most relevant to our work, \citet{butler2023ms2mol} build MS2Mol, an end-to-end language model that encodes m/z values and intensities as tokenized text input and outputs an inferred chemical formula and SMILES string in an autoregressive manner. However, their implementation is not currently available at the time of writing, preventing direct comparison. Most recently, MADGEN~\citep{wang2025madgenmassspecattendsnovo} presents a diffusion generator of chemical structures from scaffolds as a two-stage generative process, seemingly bottlenecked in terms of accuracy by scaffold prediction. 
In this paper, we improve upon this thread of end-to-end approaches by encoding inductive biases via spectral transformers and utilizing a pretraining-finetuning framework for an MS-conditioned diffusion generator. \ours has two stages like MADGEN, but is trained end-to-end during its final training step, and is heavy-atom constrained.




\begin{figure*}
    \centering
    \includegraphics[width=\linewidth]{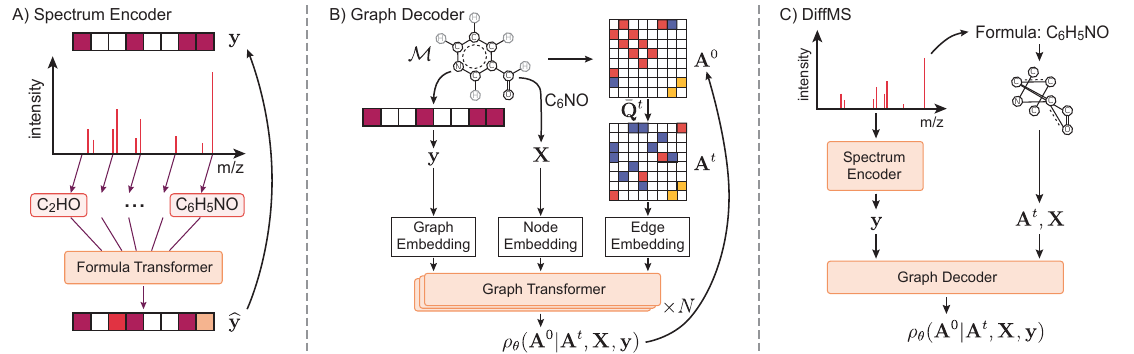}
    \caption{Model architecture of \ours. \textbf{A)}~The spectrum encoder first assigns chemical formulae to peaks in an experimental spectrum and then learns an embedding vector through a formula transformer. The encoder is pretrained to predict Morgan fingerprints~\citep{morgan1965generation} from spectra. \textbf{B)}~The graph decoder generates the target adjacency matrix by discrete diffusion conditioned on the spectrum embedding and node (atom) features. The graph decoder is pretrained with pairs of structures and fingerprints from virtual chemical libraries. We scale up the decoder pretraining to exploit the virtually-infinite number of available fingerprint-structure pairs relative to the small number of available spectrum-structure pairs, mitigating the challenge of fingerprint-to-molecule generation found non-trivial by \citet{le2020neuraldecipher}. \textbf{C)}~\ours integrates the spectrum encoder and graph decoder to generate the structure annotation as a denoising process applied to a graph with randomly generated edges. It is finetuned end-to-end on labeled molecule-spectrum data.} 
    \label{fig:model-overview}
\end{figure*}


\subsection{Diffusion generative models}
Denoising diffusion ~\citep{pmlr-v37-sohl-dickstein15, ho2020denoisingdiffusionprobabilisticmodels} has been shown to be widely effective across many tasks such as image \citep{song2020score, saharia2022photorealistic, karras2022elucidating} and text \citep{li2022diffusion, austin2023structureddenoisingdiffusionmodels} generation. More recently, diffusion has been applied to solve (bio)molecular generative tasks 
\citep{corso2023diffdockdiffusionstepstwists, Watson2023DeND, zeni2024mattergengenerativemodelinorganic}. 

Broadly speaking, diffusion models are generative models defined by a forward process that progressively adds noise to a sample $\bm{z}$ from data distribution $q(\bm{z}^0)$ such that $q(\bm{z}^T | \bm{z}^0)$ converges to a known prior distribution $p(\bm{z}^T)$ as $T \to \infty$. We additionally require that the noising process be Markovian such that $q(\bm{z}^1, \dots, \bm{z}^T | \bm{z}^0) = \prod_{t=1}^Tq(\bm{z}^t | \bm{z}^{t-1})$. Finally, we select the forward process such that we can efficiently sample from $q(\bm{z}^t | \bm{z}^0)$.

A neural network is then trained to reverse this noising process. However, instead of predicting $p_{\theta}(\bm{z}^{t-1} | \bm{z}^t)$, as long as $p_{\theta}(\bm{z}^{t-1} | \bm{z}^t) = \int q(\bm{z}^{t-1} | \bm{z}^t, \bm{z}^0)dp_{\theta}(\bm{z}^0)$ is tractable, we can train the model to directly predict the denoised sample $p_{\theta}(\bm{z}^0 | \bm{z}^t)$. To generate new samples from the model, we sample random noise from the prior distribution $p(\bm{z}^T)$, and iteratively sample from $p_{\theta}(\bm{z}^{t-1} | \bm{z}^t)$ until reaching $\bm{z}^0$.

Many works have also studied conditional generation with diffusion. Conditional diffusion models typically fall into two categories: classifier guidance \citep{dhariwal2021diffusion} and classifier-free guidance \citep{ho2022classifier}. Classifier guidance uses the gradients of the log likelihood of a classifier function $p_{\phi}(y | \bm{z}^t)$ to guide the diffusion towards samples with class $y$. On the other hand, classifier-free guidance trains the denoising network directly to generate samples conditioned on class $y$ and does not require any external classifier function. \ours falls under classifier-free guidance.   

While diffusion models were originally designed to operate in continuous spaces, recent works have adapted denoising diffusion for discrete data modalities ~\citep{austin2023structureddenoisingdiffusionmodels, lou2024discretediffusionmodelingestimating} and graph structured data ~\citep{vignac2023digress, chen2023efficientdegreeguidedgraphgeneration}, both of which are relevant for molecule generation. Here, we follow the discrete diffusion settings of ~\citet{austin2023structureddenoisingdiffusionmodels} and \citet{vignac2023digress}.

\section{Methodology}

\subsection{Formula-constrained molecular generation}


We represent structure-spectrum pairs as $\left(\mathcal{M}, \mathcal{S}\right)$, where $\mathcal{M}$ is the graph representation of the molecule with corresponding spectrum $\mathcal{S}$. The goal of \textit{de novo} generation is to reconstruct the molecular graph $\widehat{\mathcal{M}}$ from $\mathcal{S}$. Because the molecular structure is typically underspecified given the spectrum, it is more natural to formulate \textit{de novo} generation as predicting a ranked list of $k$ molecules $\widehat{\bm{\mathcal{M}}_k} = (\widehat{\mathcal{M}}_1, \dots, \widehat{\mathcal{M}}_k)$ that most closely match the given spectra.


One insight in this work is that chemical formulae represent an important physical prior that can significantly reduce the molecular search space. Formulae can be inferred from high-resolution MS1 data and isotopic traces with sufficient accuracy using tools like SIRIUS~\citep{bocker2016fragmentation} or MIST-CF~\citep{goldman2023mist-cf}, though the latter does not consider isotope distributions. To that end, we develop a formula-restricted generation using graph diffusion.  In practice, we find it sufficient to model only the heavy-atoms in the graph and infer hydrogen atom placement implicitly; thus \ours generated molecules may differ in formula from the true molecule in hydrogen atom count. 




\subsection{\ours discrete diffusion}

Let a molecular graph $\mathcal{M} = \left(\mathbf{A}, \mathbf{X}, \mathbf{y}\right)$ with one-hot encoded adjacency matrix 
$\mathbf{A} \in \{0,1\}^{n \times n \times k}$
, node features $\mathbf{X}\in \mathbb{R}^{n \times d}$ such as atom types, and graph-level structural features $\mathbf{y} \in \mathbb{R}^c$ to condition the molecule generation such as molecular fingerprint or mass spectra. Here, $n$ is the number of heavy atoms in the molecule; $k = 5$, the number of bond types (no bond, single, double, triple, and aromatic bonds); $d$, the dimension of atom features; and $c$, the dimension of the conditional features. Because we obtain atom types from the formula, we can fix $\mathbf{X}$ and generate the adjacency matrix $\mathbf{A}$ conditioned on $\mathbf{X}$ and $\mathbf{y}$.

We define a discrete diffusion process on $\mathbf{A}$. Let $\mathbf{A}^t$ denote the value of $\mathbf{A}$ at time $t$. Let $\mathbf{A}^0 = \mathbf{A}$, the true molecular adjacency matrix. At each time step, $t = 1, \dots, T$, we apply noise to each edge independently of others. Specifically, we define forward transition matrices $\left(\mathbf{Q}^1, \dots \mathbf{Q}^T\right)$ such that $\mathbf{Q}^t_{mn} = q\left(a^t = n | a^{t-1} = m \right)$. Thus:
\begin{equation}
q(\mathbf{A}^t | \mathbf{A}^{t-1}) = \mathbf{A}^{t-1}\mathbf{Q}^{t}
\end{equation}
Because the noise is described by a Markov transition process, we can directly sample $\mathbf{A}^t$ given $\mathbf{A}$ as:
\begin{equation}
q(\mathbf{A}^t | \mathbf{A}) = \mathbf{A}\mathbf{\bar{Q}}^{t}
\end{equation}
Where $\mathbf{\bar{Q}}^{t} = \mathbf{Q}^1\mathbf{Q}^2\dots \mathbf{Q}^t$. Because molecular graphs are undirected, we apply noise only to the upper triangle of $\mathbf{A}$ and symmetrize the matrix. We follow the noise schedule of \citet{vignac2023digress} and select

\begin{equation}
    \mathbf{\bar{Q}}^t = \bar{\alpha}^t\bm{I} + \bar{\beta}^t\bm{1}_k\bm{m}^\top
\end{equation}

where $\bm{m}$ is the marginal distribution of edge types in the training dataset and $\bm{m}^\top$ is the transpose of $\bm{m}$. This choice of $\mathbf{Q}^t$ converges to a prior distribution that is closer to the data than a uniform distribution over bond types, enabling easier training. We further select the cosine noise schedule proposed by \citet{nichol2021improveddenoisingdiffusionprobabilistic}:

\begin{equation}
    \bar{\alpha}^t = \cos\left(\frac{\pi(t/T + \epsilon)}{2(1+\epsilon)}\right)^2
\end{equation}

with $\bar{\beta}^t = 1 - \bar{\alpha}^t$. We then define a neural network $\phi_{\theta}$ that learns to predict the denoised adjacency matrix $\mathbf{A}^0$. Let $\mathcal{M}^t = \left(\mathbf{A}^t, \mathbf{X}, \mathbf{y}\right)$ be the noised molecule at time $t$. $\phi_{\theta}$ takes $\mathcal{M}^t$ as input and predicts probabilities $p_{\theta}(\mathbf{A}^0 |\mathcal{M}^t) = \phi_\theta(\mathcal{M}^t) \in \mathbb{R}^{n \times n \times k}$. 
i.e., it learns to denoise $\mathbf{A}^t$ conditioned on $\mathbf{X}$, $\mathbf{y}$. We optimize this network using cross-entropy loss $L$ between the true adjacency matrix $\mathbf{A}$ and the predicted probabilities $\hat{\mathbf{A}} = \phi_\theta(\mathcal{M}^t)$:
\begin{equation}
L(\mathbf{A}, \hat{\mathbf{A}}) = \sum_{1 \leq i < j \leq n}\text{CE}\left(a_{ij}, \hat{a}_{ij}\right)
\end{equation}

To sample new graphs, we need to compute $p_{\theta}\left(\mathbf{A}^{t-1} | \mathcal{M}^t\right)$. We 
 do so by marginalizing over the network predictions:

\begin{equation}
    p_{\theta}\left(a_{ij}^{t-1} | \mathcal{M}^t\right) = \sum_{a_k}p_{\theta}\left(a_{ij}^{t-1} | a_{ij} = a_k, \mathcal{M}^t\right)p_{\theta}(a_k)
\end{equation}

using $p_{\theta}\left(a_{ij}^{t-1} | a_{ij} = a_k, \mathcal{M}^t\right) = q\left(a_{ij}^{t-1} | a_{ij}=a_k, a_{ij}^t\right)$ if $q\left(a_{ij}^t | a_{ij}=a_k\right) > 0$, otherwise $0$. We can then generate new graphs by sampling an initial $\mathbf{A}^T \sim \bm{m}$ and iteratively sampling from $p_{\theta}\left(\mathbf{A}^{t-1} | \mathcal{M}^t\right)$ until we obtain $\mathbf{A}^0$.

\subsection{Model parametrization and pretraining}

We use an encoder-decoder architecture to enable separate pretraining for the encoder and decoder before finetuning the end-to-end generative model. Specifically, a spectrum encoder infers  structural information from $\mathcal{S}$ and the encoder embeddings are used as the structural condition $\mathbf{y}$ for the graph diffusion decoder (Fig.~\ref{fig:model-overview}). 

For the encoder module, we use the MIST formula transformer of \citet{goldman2023mist}. The encoder treats a spectrum as a set of (m/z, intensity) peaks. It embeds each peak using a predicted chemical formula assignment from SIRIUS and applies a set transformer that implicitly models pairwise neutral losses between fragments. We extract the final embedding corresponding to the precursor peak as the structural condition $\mathbf{y}$ for the diffusion decoder.  

We pretrain our encoder on the same datasets used for finetuning (i.e., NPLIB1 (CANOPUS) or MassSpecGym) but now train the encoder to predict molecular fingerprints. We find that this pretraining enables the encoder to extract implicit structural information from the spectra and ensures that the encoder learns physically meaningful representations. We provide an ablation of the encoder pretraining in Sec. \ref{sec:ablations}.

For the decoder network $\phi_\theta$ that predicts the denoised adjacency matrix, we use a Graph Transformer \citep{dwivedi2021generalizationtransformernetworksgraphs}. Specifically, we use separate MLPs to encode edge features $\mathbf{A}^t$, node features $\mathbf{X}$, and structural condition $\mathbf{y}$. We then apply several Graph Transformer layers before using an MLP to predict the denoised adjacency matrix $\hat{\mathbf{A}}$.



We pretrain our diffusion decoder on a dataset of fingerprint-molecule pairs. Instead of using the spectrum encoder embeddings as the structural condition $\mathbf{y}$, we directly use the molecular fingerprint to condition the molecule generation. This is closely aligned with the mass spectra \textit{de novo} generation task, as the decoder learns to generate molecules subject to strong structural constraints. 
Fingerprint-molecule datasets are essentially infinite in size, providing a promising path forward to further improve model performance by increasing the pretraining dataset size. To this end, we build a pretraining dataset consisting of 2.8M fingerprint-molecule pairs sampled from  DSSTox \citep{CCTE2019}, HMDB \citep{hmdb}, COCONUT \citep{Sorokina_Merseburger_Rajan_Yirik_Steinbeck_2021}, and MOSES \citep{polykovskiy2020molecularsetsmosesbenchmarking} datasets. Critically, we remove \emph{all} NPLIB1 and MassSpecGym test and validation molecules from our decoder pretraining dataset so that our evaluation on the end-to-end generation task represents a setting where the model is generating truly novel structures. \citet{bushuiev2024massspecgymbenchmarkdiscoveryidentification} provide their own dataset of 4M molecules, but use a different exclusion criteria to prevent data leakage. We provide an ablation and analysis of performance scaling with respect to pretraining dataset size in Sec. \ref{sec:ablations}.


\begin{table*}[t]
\caption{\textit{De novo} structural elucidation performance on NPLIB1 \citep{duhrkop2021canopus} and MassSpecGym~\citep{bushuiev2024massspecgymbenchmarkdiscoveryidentification} datasets. The best performing model for each metric is \textbf{bold} and the second best is \underline{underlined}. $\ddag$ indicates results reproduced from MassSpecGym.
$*$ indicates our implementations of baseline approaches. Methods are approximately ordered by performance.
}
\vspace{-0.05in}
\label{table:main}
\begin{center}
{
\begin{tabular}{lcccccc}
\toprule
& \multicolumn{3}{c}{Top-1} & \multicolumn{3}{c}{Top-10} \\
\cmidrule(lr){2-4}
\cmidrule(lr){5-7}
Model & Accuracy $\uparrow$ & MCES $\downarrow$ & Tanimoto $\uparrow$ & Accuracy $\uparrow$ & MCES $\downarrow$ & Tanimoto $\uparrow$ \\
\midrule
\midrule
& \multicolumn{4}{c}{NPLIB1} & & \\
\midrule
Spec2Mol$^*$ & 0.00\% & 27.82 & 0.12 & 0.00\% & 23.13 & 0.16 \\
MADGEN & 2.10\% & 20.56 & 0.22 & 2.39\% & 12.64 & 0.27 \\
MIST + Neuraldecipher$^*$ & 2.32\% & \underline{12.11} & \textbf{0.35} & 6.11\% & \underline{9.91} & 0.43 \\
MIST + MSNovelist$^*$ & \underline{5.40\%} & 14.52 & 0.34 & \underline{11.04}\% & 10.23 & \underline{0.44} \\
DiffMS & \textbf{8.34\%} & \textbf{11.95} & \textbf{0.35} & \textbf{15.44\%} & \textbf{9.23} & \textbf{0.47} \\
\midrule
& \multicolumn{4}{c}{MassSpecGym} & & \\
\midrule
SMILES Transformer$^\ddag$ & 0.00\% &  79.39 & 0.03 & 0.00\% & 52.13 & 0.10 \\
MIST + MSNovelist$^*$ & 0.00\% & 45.55 & 0.06 & 0.00\% & 30.13 & 0.15 \\
SELFIES Transformer$^\ddag$ & 0.00\% & 38.88 & 0.08 & 0.00\% & 26.87 & 0.13 \\
Spec2Mol$^*$ & 0.00\% & 37.76 & 0.12 & 0.00\% & 29.40 & 0.16\\
MIST + Neuraldecipher$^*$ & 0.00\% & 33.19 & 0.14 & 0.00\% & 31.89 & 0.16 \\
Random Generation$^\ddag$ & 0.00\% & \underline{21.11} & 0.08 & 0.00\% & 18.26 & 0.11 \\
MADGEN & \underline{1.31\%} & 27.47 & \underline{0.20} & \underline{1.54}\% & \underline{16.84} & \underline{0.26} \\
DiffMS & \textbf{2.30\%} & \textbf{18.45} & \textbf{0.28} & \textbf{4.25\%} & \textbf{14.73} & \textbf{0.39} \\
\bottomrule
\end{tabular}
}
\end{center}
\vskip -0.1in
\end{table*}

\section{Experiments}
\subsection{Evaluation metrics}
We adopt the \emph{de novo} generation metrics from ~\citet{bushuiev2024massspecgymbenchmarkdiscoveryidentification}:
\begin{itemize}[noitemsep,topsep=0pt] 
    \item Top-$k$ accuracy: measures whether the true molecule is in the top-$k$ model predictions.
    \item Top-$k$ maximum Tanimoto similarity: the structural similarity of the closest molecule to the true molecule in the top-$k$ predictions 
    \item Top-$k$ minimum MCES (maximum common edge subgraph): the graph edit distance of the closest molecule to the true molecule in the top-$k$ predictions using the distance metric proposed by \citet{Kretschmer2023.03.27.534311}. 
\end{itemize}


We report metrics for $k=1, 10$ with additional results in Appendix \ref{appendix:ablations}. 
To obtain a ranked list of \ours predictions, we sample 100 molecules for each spectrum, remove invalid or disconnected molecules, and identify the top-$k$ molecules based on frequency. This post-processing is also applied to baseline methods for fairest comparison.
\subsection{Datasets and baselines}

We evaluate \ours on two common open-source \textit{de novo} generation benchmark datasets, NPLIB1 \citep{Dührkop2021} and MassSpecGym \citep{bushuiev2024massspecgymbenchmarkdiscoveryidentification}. The NPLIB1 dataset is the subset of GNPS data used to train the CANOPUS tool; this term is used to disambiguate the data from the method. In order to have a fair evaluation of all methods considered, we re-implement several baseline methods to be trained only on these datasets, with modifications to the codebase if they did not have a working open-source implementation. While some papers have historically also benchmarked on the NIST20 or NIST23 datasets~\citep{nist_database}, this dataset is not publicly available without purchase of a license. 

MSNovelist~\citep{stravs2022msnovelist} builds a fingerprint-to-SMILES LSTM decoder to predict SMILES strings from the SIRIUS-generated CSI-FingerID fingerprint. The original implementation of MSNovelist is not readily retrainable, and furthermore relies on the closed-source CSI-FingerID fingerprint. The recently developed MIST model offers an open-source replacement with reported comparative performance to CSI-FingerID~\citep{goldman2023mist}. 
Accordingly, we re-implement a baseline model that retains the main contributions of MSNovelist, adopting the code from \citet{Zhao2024-ew}, with a 4096-bit Morgan fingerprint spectral encoder using MIST alongside a formula-guided fingerprint-to-SMILES LSTM. 
This fingerprint-to-SMILES decoder is trained on the same 2.8M dataset used to pretrain our diffusion decoder; therefore, unlike the original MSNovelist implementation, both the spectral encoder and LSTM decoder \emph{never} see 
any test structures. We use the ranking methodology from MSNovelist, wherein beam search (with a width of 100) and subsequently computed log-likelihoods are used for ranking. 
Similarly, Spec2Mol~\citep{litsa2023spec2mol} 
was also retrained on the NPLIB1 and MassSpecGym datasets for fair evaluation, with only one spectral channel instead of four used for training, to alleviate restrictions on collision energy or adduct. The same ranking for candidate molecules as used for \ours is applied.


We also introduce a new baseline method, MIST + Neuraldecipher, that replaces the diffusion decoder in \ours with Neuraldecipher. Neuraldecipher encodes a molecule into a CDDD representation \citep{C8SC04175J}, and uses a pretrained LSTM decoder to reconstruct the SMILES string. Similar to \ours, we pretrain the MIST encoder on spectrum-to-fingerprint predictions, and we pretrain Neuraldecipher on fingerprint-to-molecule generation. Since MIST + Neuraldecipher uses the same pretraining-finetuning approach as \ours, this new baseline additionally serves as an empirical justification for our graph diffusion decoder over an LSTM-based approach.

Finally, we include a comparison to MADGEN~\citep{wang2025madgenmassspecattendsnovo}. The MADGEN$_{\text{Oracle}}$ entry in \citet{wang2025madgenmassspecattendsnovo} feeds in the ground-truth scaffold which 
does not fall within the setting of complete \emph{de novo} generation, and is thus not included in our evaluation. Because MADGEN uses RDKFingerprints for evaluation, as opposed to the traditional Morgan fingerprint, we exclude their Tanimoto similarities.


\subsection{Results}
As seen in Table \ref{table:main}, \ours outperforms baseline methods on both datasets, including more than doubling the accuracy on MassSpecGym compared to the next best method, MADGEN. While there are several baseline methods that achieve non-zero prediction accuracy on NPLIB1, only MADGEN and \ours generate any correct structures on MassSpecGym. NPLIB1 is inherently a less challenging dataset than MassSpecGym; given the lack of a scaffold-based split, the CANOPUS test set contains many molecules that are nearly identical (Tanimoto similarity $>$ 0.85) to molecules in the train set \citep{bushuiev2024massspecgymbenchmarkdiscoveryidentification}. This also explains the competitive performance of MIST + Neuraldecipher and MIST + MSNovelist, which both benefit from their ability to pretrain on these highly similar structures and learn to generate realistic structures as SMILES strings; the MSNovelist generation even more so given its formula-aware decoder. In contrast, MassSpecGym ensures that no molecules in the test set have an MCES $<$ 10 compared to any training molecule. As such, MassSpecGym evaluation represents a more challenging and more realistic out of distribution \textit{de novo} generation setting, which illustrates the robust performance of \ours across all evaluation metrics.
Examples of \ours-generated sampled are shown in Figure~\ref{fig:example_preds_main} and Appendix \ref{appendix:molecules}. In Appendix \ref{appendix:eval}, we show that even in cases where \ours does not recover the correct structure, it is consistently able to generate ``close match`` structures that are still useful to domain experts.

\begin{figure*}[h]
\centering

\subfigure{\includegraphics[width=1.0\textwidth]{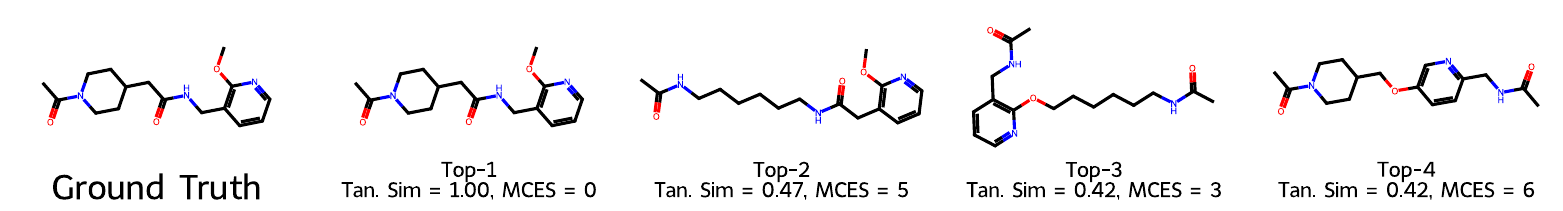}}

\subfigure{\includegraphics[width=1.0\textwidth]{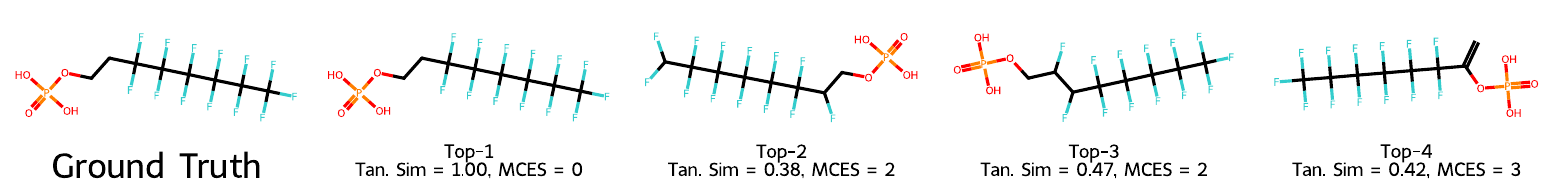}}

\subfigure{\includegraphics[width=1.0\textwidth]{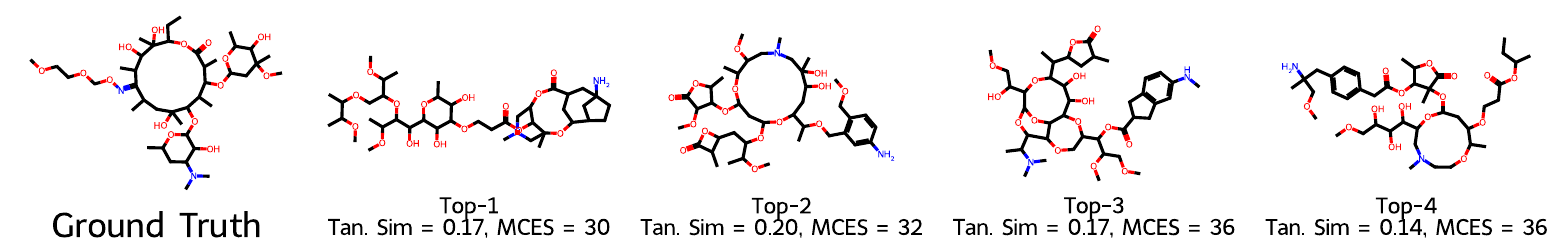}}

\subfigure{\includegraphics[width=1.0\textwidth]{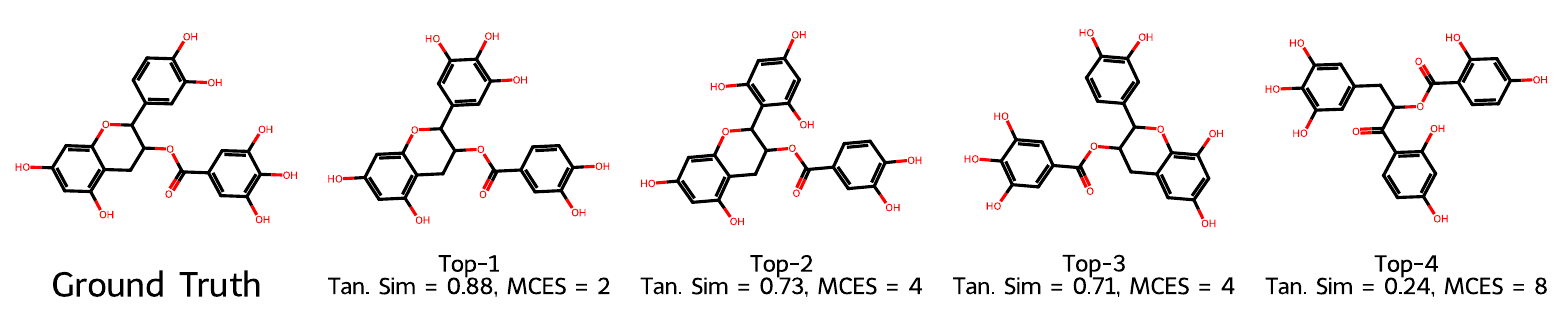}}

\vspace{-0.1in}
\caption{Ground truth molecules (left column) and \ours predictions (right columns) on test samples from the MassSpecGym dataset~\citep{bushuiev2024massspecgymbenchmarkdiscoveryidentification}. Tanimoto similarity and MCES metrics listed for each top-$k$ prediction. From top to bottom, the spectra IDs are MassSpecGymID0205184, MassSpecGymID0052933, MassSpecGymID0382596, and MassSpecGymID0152454. The top two rows show cases where \ours successfully reconstructs the true molecule in the top-1 prediction. In the bottom two rows, \ours does not reconstruct the correct molecule. Additional examples can be found in Appendix \ref{appendix:molecules}.}
\label{fig:example_preds_main}
\end{figure*}

\subsection{Pretraining Ablations} \label{sec:ablations}

To highlight the performance gains from pretraining the \ours encoder and decoder, we provide several ablations. Note that comparisons in Table~\ref{table:main} to MIST + Neuraldecipher and MIST + MSNovelist already serve as empirical justification for \ours' discrete graph decoder.

\begin{table}
\renewcommand{\arraystretch}{1.1}
\vspace{-0.2in}
\caption{\ours performance on NPLIB1 with and without pretraining the MIST encoder on the spectrum-to-fingerprint task. The best performing model for each metric is \textbf{bold}.}
\label{table:ablate_encoder}
\vspace{-0.05in}
\begin{center}
\begin{tabular}{c|ccc}
\toprule
  Pretrain? & Accuracy $\uparrow$ & MCES $\downarrow$ & Tanimoto $\uparrow$ \\
 \midrule
 \midrule 
\multicolumn{4}{c}{Top-1} \\
\midrule
\xmark & 4.36\% & 12.34 & 0.31 \\
\cmark & \textbf{8.34\%} & \textbf{11.95} & \textbf{0.35} \\
\midrule 
\multicolumn{4}{c}{Top-10} \\
\midrule
\xmark & 11.46\% & 9.31 & 0.44\\
\cmark & \textbf{15.44\%} & \textbf{9.23} & \textbf{0.47} \\
\bottomrule
\end{tabular}
\end{center}
\vspace{-0.2in}
\end{table}

\begin{table*}[th!]
\caption{DiffMS \textit{de novo} structural elucidation performance on NPLIB1 \citep{duhrkop2021canopus} and MassSpecGym~\citep{bushuiev2024massspecgymbenchmarkdiscoveryidentification} datasets using MIST-CF annotated formulae and ground truth formulae. The best performing model for each metric is \textbf{bold}.
}
\label{table:formula_study}
\begin{center}
{
\begin{tabular}{lcccccc}
\toprule
& \multicolumn{3}{c}{Top-1} & \multicolumn{3}{c}{Top-10} \\
\cmidrule(lr){2-4}
\cmidrule(lr){5-7}
Formulae & Accuracy $\uparrow$ & MCES $\downarrow$ & Tanimoto $\uparrow$ & Accuracy $\uparrow$ & MCES $\downarrow$ & Tanimoto $\uparrow$ \\
\midrule
\midrule
& \multicolumn{4}{c}{NPLIB1} & & \\
\midrule
MIST-CF Formulae & 7.03\% & \textbf{11.81} & \textbf{0.36} & 14.98\% & 9.39 & \textbf{0.48} \\
True Formulae & \textbf{8.34\%} & 11.95 & 0.35 & \textbf{15.44\%} & \textbf{9.23} & 0.47 \\
\midrule
& \multicolumn{4}{c}{MassSpecGym} & & \\
\midrule
MIST-CF Formulae & 1.86\% & \textbf{17.83} & 0.27 & 4.10\% & \textbf{13.71} & \textbf{0.40} \\
True Formulae & \textbf{2.30\%} & 18.45 & \textbf{0.28} & \textbf{4.25\%} & 14.73 & 0.39 \\
\bottomrule
\end{tabular}
}
\end{center}
\vskip -0.1in
\end{table*}

\textbf{Encoder Pretraining Ablation.} We train \ours without pretraining the MIST encoder on the spectra-to-fingerprint task. As demonstrated in Table \ref{table:ablate_encoder}, encoder pretraining provides significant performance gains on the NPLIB1 dataset, nearly doubling the top-1 accuracy. Additionally, we see 
that even without pretraining the encoder, \ours generates realistic, plausible structures as indicated by the MCES and Tanimoto metrics. Nonetheless, the encoder pretraining improves the ability of the decoder to condition the diffusion on the spectra and obtain an exact match.  

\begin{figure}
    \centering
        \vspace{-0.1in}
    \includegraphics[width=1.0\linewidth]{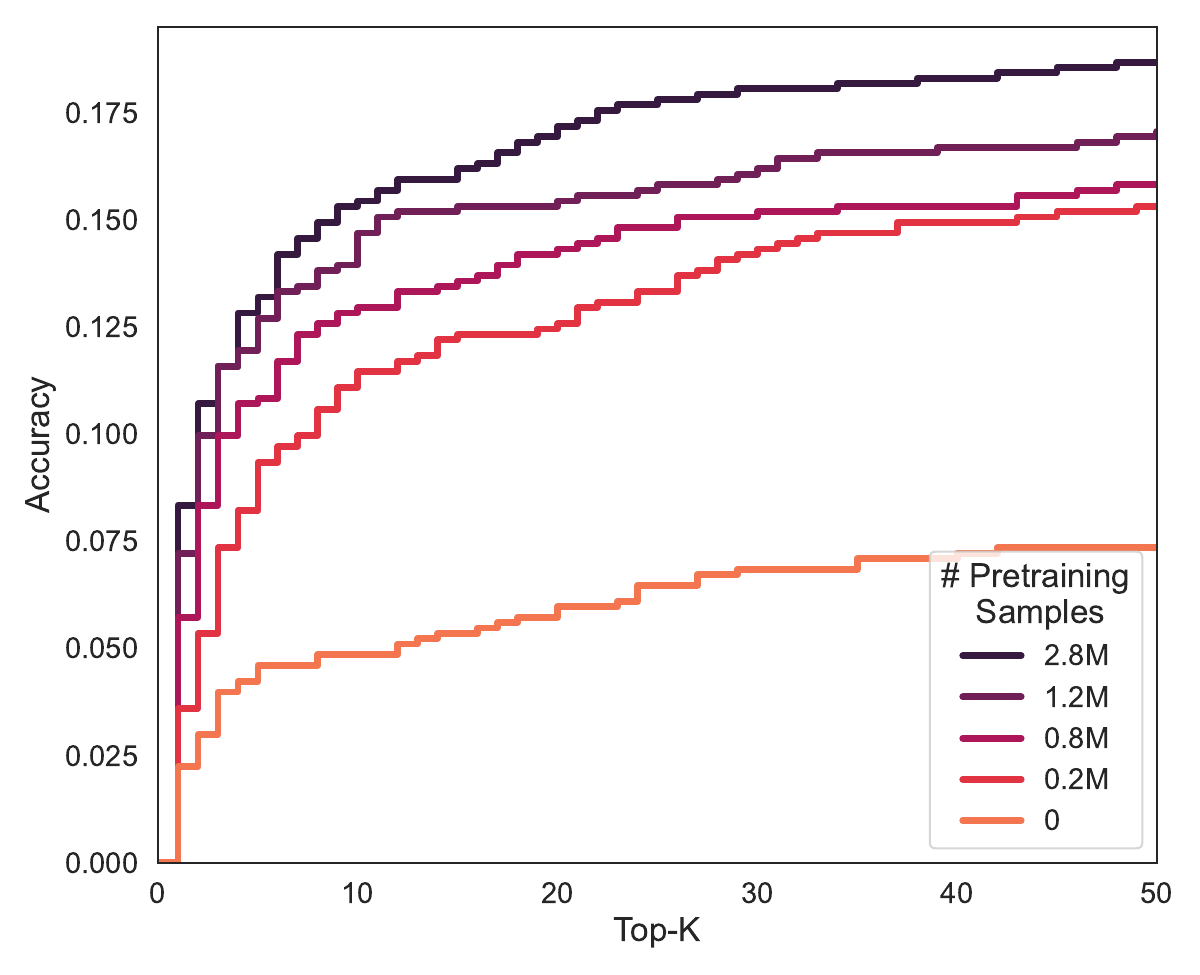}
    \vspace{-0.25in}
    \caption{NPLIB1 top-$k$ accuracy for \ours pretrained on increasingly large fingerprint-to-molecule datasets. Additional metrics available in Table \ref{table:ablate_decoder} in the Appendix.}
    \label{fig:ablate_decoder}
    \vspace{-0.1in}
\end{figure}

\textbf{Decoder Pretraining Ablation.} We train \ours with increasing decoder pretraining dataset size, starting from 0 molecules (i.e., no pretraining) up to the full pretraining dataset of 2.8M molecules. As shown in Fig.~\ref{fig:ablate_decoder}, any amount of decoder pretraining offers a significant increase in performance. Additionally, we observe good performance scaling with increasing pretraining dataset size. Since fingerprint-to-molecule datasets are essentially infinite in size, this provides an avenue to continue scaling \ours' performance by building even larger and more chemically comprehensive pretraining datasets.

Additional results and figures for encoder and decoder pretraining ablations can be found in Appendix \ref{appendix:ablations}.

\subsection{Formula Inference Ablation}
\label{sec:formula_ablation}

In many real-world elucidation settings, chemists may know the true chemical formula of the target compound \textit{a priori} or be able to determine the true formula using auxiliary methods. However, chemical formulae can also be predicted from the spectrum with high accuracy by out-of-the-box formula annotation tools~\citep{goldman2023mist-cf, Xing2023-buddy, bocker2016fragmentation}. In this section, we broaden the structural elucidation challenge and investigate the ability of DiffMS to rely on MIST-CF~\citep{goldman2023mist-cf} formula predictions to test its performance in settings where the true chemical formula is unknown.

For each spectrum, we predict the top 5 most likely formulae using MIST-CF. We then generate candidate structures for each of the predicted formulae. To have a fair comparison, we still generate 100 total molecules, split across the 5 predicted formulae. As shown in Table~\ref{table:formula_study}, DiffMS still has strong performance even when relying on formula annotation tools to supply the formula. While the elucidation accuracy is slightly lower, the MCES and Tanimoto metrics are actually better in some cases using the MIST-CF predicted formulae. Intuitively, sampling molecules with different formulae gives us higher diversity and thus a better chance of getting a ``close'' structure.  

\section{Conclusion}

In this work, we propose \ours, a conditional molecule generative model with formula constraints for structural elucidation from mass spectra. We develop a pretraining-finetuning framework for separate spectra encoder and graph diffusion decoders that makes use of extensive fingerprint-molecule datasets and ensures the spectrum encoder learns to extract physically meaningful representations from mass spectra. We show that \ours achieves state-of-the-art results across common \textit{de novo} generation benchmarks, and provide several ablations to demonstrate the effectiveness of our contributions and the potential to further improve performance by scaling pretraining.



\section*{Acknowledgments}
This work was partly sponsored by DSO National Laboratories in Singapore (to C.W.C.), the MIT Summer Research Program (MSRP), National Science Foundation under grant IIS-2243850 (to S.J.), and ARPA-H under grant 1AY1AX000053 (to S.J.).

\section*{Impact Statement}
The advancement of computational tools for structure elucidation will aid in the identification of unknown molecules, including metabolites as biomarkers for diagnostic applications or improving understanding of biology. There are many potential beneficial societal consequences of our work and very few potential negative ones, none of which warrant elaboration here.





\bibliography{example_paper}
\bibliographystyle{icml2025}

\newpage
\appendix
\onecolumn

\begin{table*}[t]
\caption{Additional evaluation of molecule validity, and percentage above domain-expert-defined Tanimoto thresholds on NPLIB1 ~\citep{duhrkop2021canopus} and MassSpecGym~\citep{bushuiev2024massspecgymbenchmarkdiscoveryidentification} \textit{de novo} generation datasets. The best performing model for each metric is \textbf{bold} and the second best is \underline{underlined}. Definitions of meaningful match (Tanimoto similarity $\geq0.4$) and close match (Tanimoto similarity $\geq 0.675$) are taken from \citet{butler2023ms2mol}.
}

\label{table:addl_main}
\begin{center}
\begin{sc}
\resizebox{\linewidth}{!}{
\begin{tabular}{lccccc}
\toprule
& Overall & \multicolumn{2}{c}{Top-1} & \multicolumn{2}{c}{Top-10} \\
\cmidrule(lr){2-2}
\cmidrule(lr){3-4}
\cmidrule(lr){5-6}
Model & \% Valid $\uparrow$ & \% Meaningful match $\uparrow$ & \% Close match $\uparrow$  & \% Meaningful match $\uparrow$ & \% Close match $\uparrow$ \\
\midrule
\midrule
& & \multicolumn{3}{c}{NPLIB1}  & \\
\midrule
Spec2Mol & 66.5\% & 0.00\% & 0.00\%& 0.00\%&0.00\% \\
MIST + Neuraldecipher & 91.11\% & \underline{29.30}\% & 7.33\% & 41.39\% &  12.82\% \\
MIST + MSNovelist  & \underline{98.60\%} & \textbf{32.90\%} &  \underline{11.78\%} &\underline{44.79\%} & \underline{19.02\%}  \\
DiffMS  & \textbf{100.0\%} & 27.40\% & \textbf{12.83\%} & \textbf{46.45\%} & \textbf{22.04\%}  \\
\midrule
& & \multicolumn{3}{c}{MassSpecGym} & \\
\midrule
Spec2Mol & 68.5\% & 0.0\% &  0.0\% &   0.0\% &  0.0\% \\
MIST + Neuraldecipher & 81.78\% & 0.29\% & \underline{0.01\%} & 0.39\% & \underline{0.09\%} \\
MIST + MSNovelist & \underline{98.58\%} & \underline{0.66\%} & 0.00\% & \underline{1.92\%} & 0.00\%  \\

DiffMS & \textbf{100.0\%}  & \textbf{12.41\%} & \textbf{3.78\%} & \textbf{32.47\%} & \textbf{6.73\%}  \\
\bottomrule
\end{tabular}
}
\end{sc}
\end{center}
\vskip -0.1in
\end{table*}

\section{Experimental Details} \label{appendix:exp_details}

For node features $\mathbf{X}$, we use a one-hot encoding of atom types, $\mathbf{X} \in \mathbb{R}^{n \times d}$, where $d$ is the number of different atom types in the dataset.

For pretraining the decoder, we use 2048-bit Morgan fingerprints with radius 2 for the structural conditioning $\mathbf{y}\in\mathbb{R}^{2048}$. We use the same training objective as the end-to-end finetuning, i.e., minimizing the cross-entropy loss between the denoised adjacency matrix $\mathbf{\hat{A}}$ and the true adjacency matrix, $\mathbf{\hat{A}}$. We build a decoder pretraining datset consisting of 2.8M fingerprint-molecule pairs sampled from  DSSTox \citep{CCTE2019}, HMDB \citep{hmdb}, COCONUT \citep{Sorokina_Merseburger_Rajan_Yirik_Steinbeck_2021}, and MOSES \citep{polykovskiy2020molecularsetsmosesbenchmarking} datasets. We pretrain the decoder for 100 epochs using the AdamW optimizer~\citep{loshchilov2017sgdrstochasticgradientdescent} and a cosine annealing learning rate scheduler~\citep{loshchilov2019decoupledweightdecayregularization}.

We pretrain the encoder on the same dataset used for finetuning (i.e. NPLIB1, MassSpecGym), which are orders of magnitude smaller than the decoder pretraining dataset. For encoder pretraining, we use the multi-objective loss settings of \citet{goldman2023mist}. We pretrain the encoder for 100 epochs using the RAdam optimizer ~\citep{liu2021varianceadaptivelearningrate}.

We finetune the end-to-end model using cross-entropy loss and no auxiliary training objectives, i.e. only the denoising diffusion objective. We use the AdamW optimizer with cosine annealing learning rate schedule for finetuning. We finetune DiffMS for 50 epochs on NPLIB1 and 15 epochs on MassSpecGym.

DiffMS is a relatively lightweight model, and all experiments were run on NVIDIA 2080ti GPUs with 12 GB of memory. On these GPUs, finetuning DiffMS takes 1.45 minutes per epoch on CANOPUS and 46 minutes per epoch on MassSpecGym. It takes 4 minutes on average to generate 100 samples from DiffMS.

\section{Additional Results} \label{appendix:eval}
As an addendum to the evaluations in Table~\ref{table:main}, we provide some additional metrics to further contextualize DiffMS performance. Firstly, we evaluate the percentage of model samples that correspond to valid molecules. Additionally, we adopt the domain-expert thresholds put forth by MS2Mol \citep{butler2023ms2mol}, where we evaluate whether candidate molecules were a ``meaningful'' match in structural similarity, having a Tanimoto similarity of 0.4 or greater; or a ``close match'' in structural similarity, having a Tanimoto similarity of 0.675 or greater. We omit MADGEN and the baseline methods from~\citet{bushuiev2024massspecgymbenchmarkdiscoveryidentification} as they do not report these metrics.

As shown in Table~\ref{table:addl_main}, 100\% of DiffMS samples are valid molecules. This is directly enforced because of our graph-based representation. In contrast, SMILES strings generated by baseline methods may not correspond to a valid structure. We find that DiffMS consistently achieves higher meaningful and close match rates than baseline methods. Impressively, DiffMS achieves over 32 times more meaningful matches in the top-10 predictions than the next best baseline on MassSpecGym. These results show that while generating exact matches continues to be a challenging task for \textit{de novo} structural elucidation, DiffMS is able to generate meaningful structural matches at a high rate.

\begin{figure}[t]
\centering
\subfigure{\includegraphics[width=0.49\textwidth]{figures/decoder_ablate_acc.pdf}}
\subfigure{\includegraphics[width=0.49\textwidth]{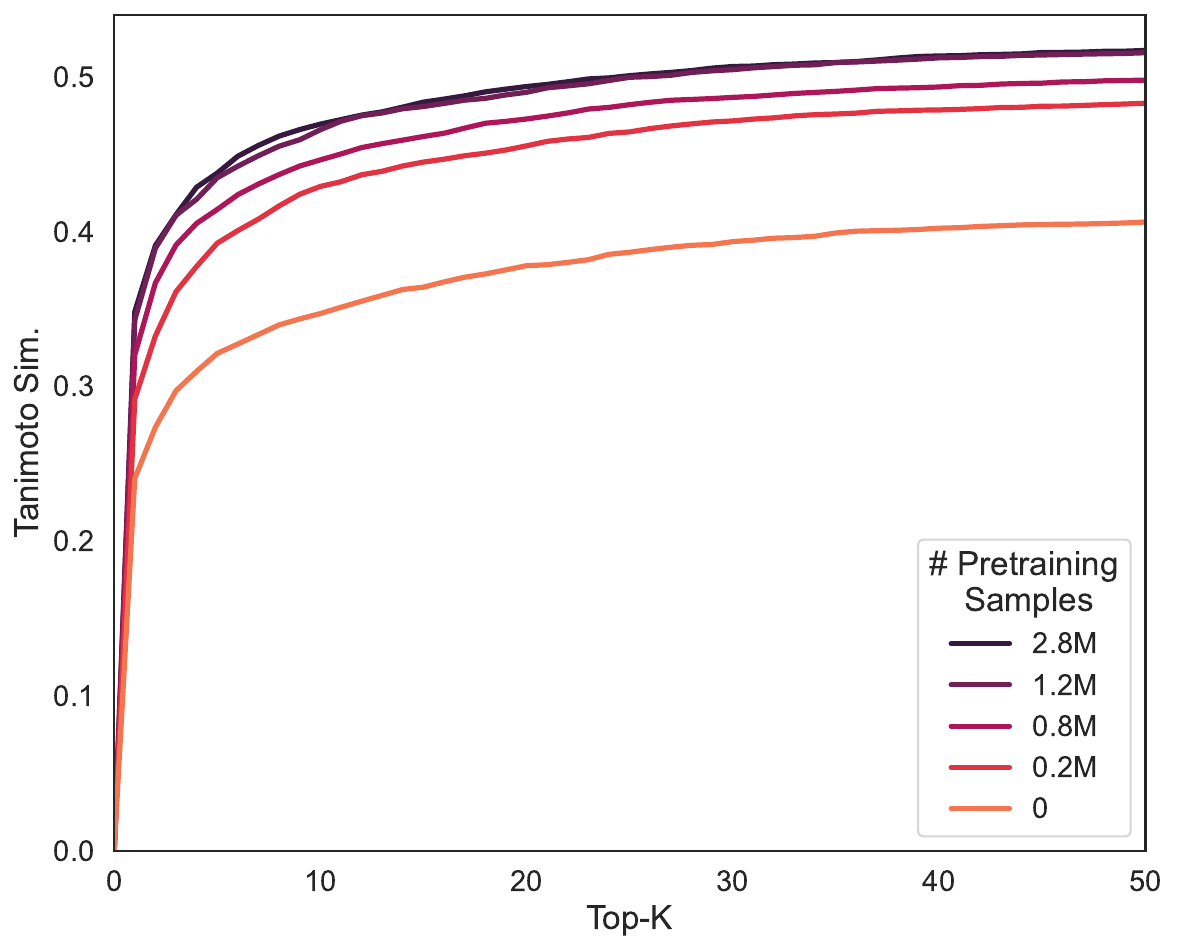}}
\vspace{-0.1in}
\caption{Annotation accuracy (left) and Tanimoto similarity (right) on the NPLIB1 dataset for \ours pretrained on increasingly large pretraining datasets.}
\label{figure:appendix_ablate_encoder}
\vspace{-0.2in}
\end{figure}

\begin{table*}[t]
\caption{DiffMS performance on NPLIB1 for \ours pretrained on increasingly large fingerprint-to-molecule datasets.
The best performing model for each metric is \textbf{bold} and second best is \underline{underlined}.}
\label{table:ablate_decoder}
\vskip 0.1in
\begin{center}
\begin{sc}
\begin{tabular}{c|cccccc}
\toprule
\multirow{2}{*}{\raisebox{-0.7\height}{\shortstack{\# Pretraining \\ Structures}}} 
& \multicolumn{3}{c}{Top-1} & \multicolumn{3}{c}{Top-10} \\
\cmidrule(lr){2-4}
\cmidrule(lr){5-7}
& Accuracy $\uparrow$ & MCES $\downarrow$ & Tanimoto $\uparrow$ & Accuracy $\uparrow$ & MCES $\downarrow$ & Tanimoto $\uparrow$ \\
\midrule
0 & 2.22\% & 15.37 & 0.22 & 4.86\% & 12.06 & 0.34 \\
0.2M & 3.61\% & 13.22 & 0.28 & 10.71\% & 9.85 & 0.41\\
0.8M & 5.60\% & 13.02 & 0.30 & 12.70\% & 9.86 & 0.44 \\
1.2M & \underline{7.22\%} & \textbf{11.63} & \underline{0.33} & \underline{14.69\%} & \textbf{9.23} & \underline{0.43} \\
2.8M & \textbf{8.34\%} & \underline{11.95} & \textbf{0.35} & \textbf{15.44\%} & \textbf{9.23} & \textbf{0.47} \\
\bottomrule
\end{tabular}
\end{sc}
\end{center}
\vskip -0.1in
\end{table*}

\section{Ablations}
\label{appendix:ablations}

\subsection{Additional Pretraining Ablation Results}
\begin{figure}[h]
\centering
\subfigure{\includegraphics[width=0.49\textwidth]{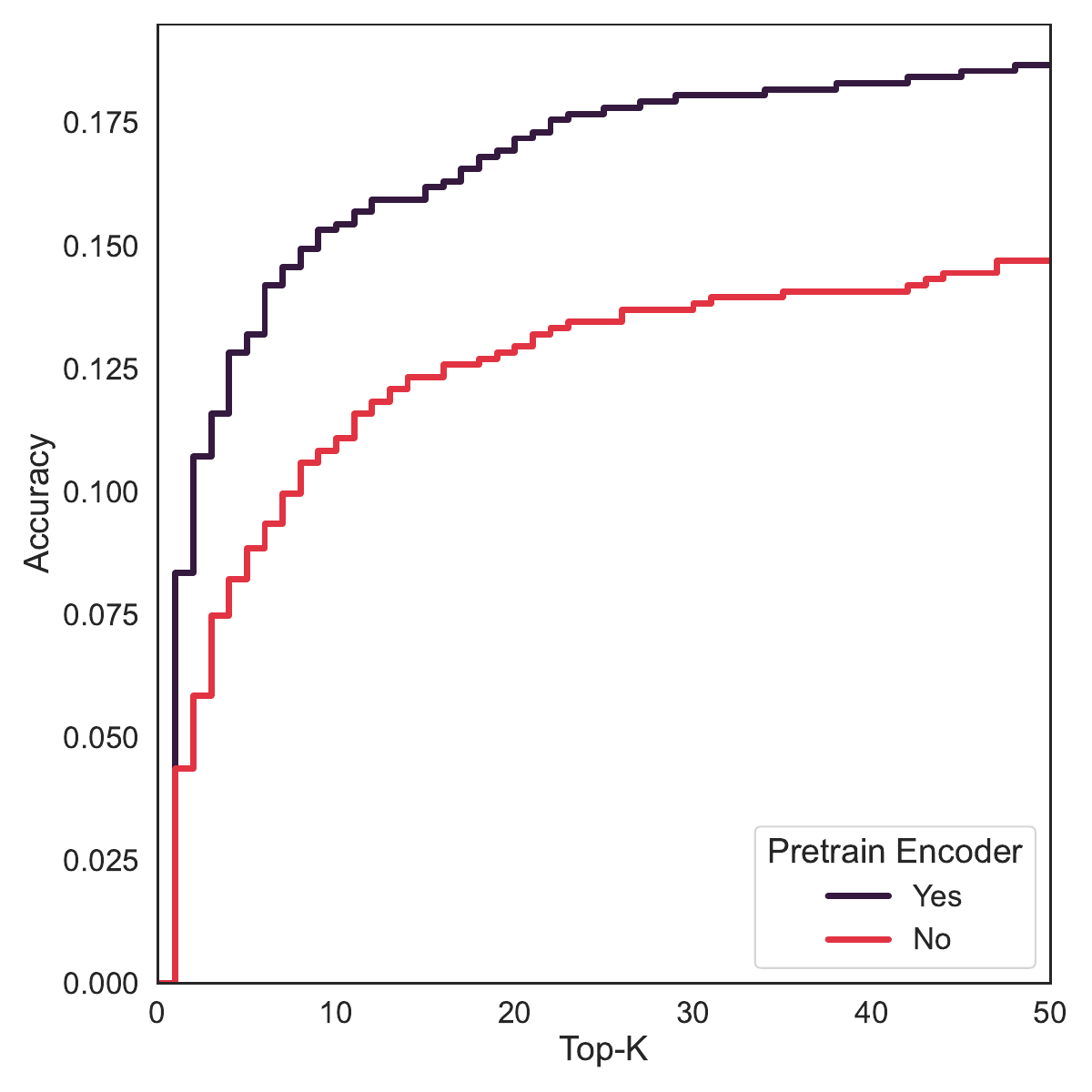}}
\subfigure{\includegraphics[width=0.49\textwidth]{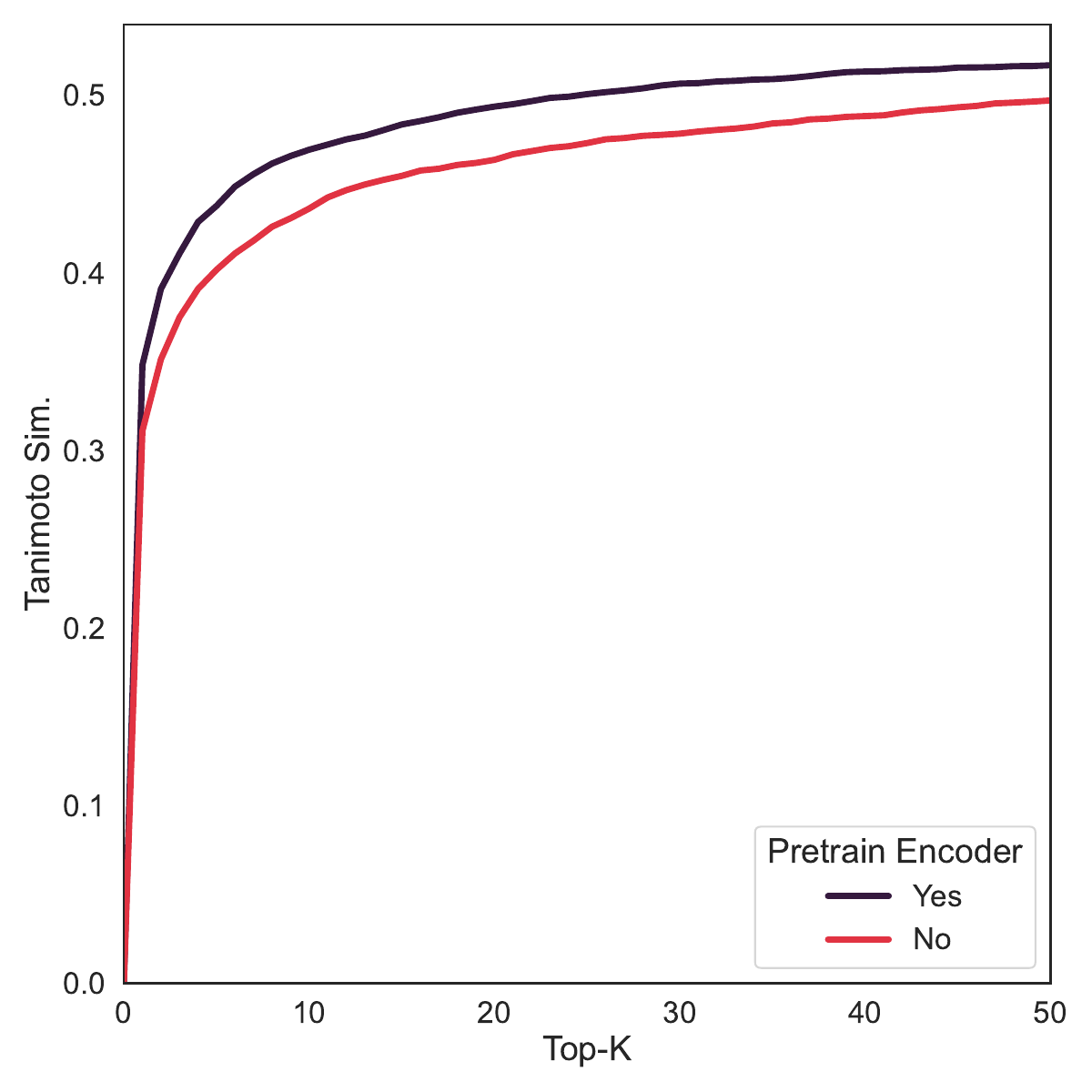}}
\vspace{-0.1in}
\caption{Annotation accuracy (left) and Tanimoto similarity (right) on the NPLIB1 dataset for \ours with and without encoder pretraining.}
\label{figure:appendix_ablate_decoder}
\end{figure}

In this section, we provide additional results for the ablation studies in Sec.~\ref{sec:ablations}. Table~\ref{table:ablate_decoder} and Fig.~\ref{figure:appendix_ablate_decoder} demonstrate DiffMS' performance scaling with respect to increasingly large decoder pretraining datasets, and Fig~\ref{figure:appendix_ablate_encoder} shows the impact of pretraining the spectra encoder.

\subsection{Prior Distribution Ablations}
In this section we provide an additional ablation study to justify the choice of the marginal prior distribution. Specifically, we compare with two alternative prior distributions: the ``empty'' distribution, consisting of no bonds, and the ``fully connected'' distribution, consisting of all single bonds. As shown in Table~\ref{table:prior_ablate}, the marginal distribution performs best, though the empty distribution is not far behind. Intuitively, the empty distribution is close to the marginal distribution as molecular graphs are typically very sparse. These results support our intuitions that having a prior distribution that is closer to the data distribution results in better performance.

\begin{table*}[h]
\caption{DiffMS performance on NPLIB1 with different prior distributions.
The best performing model for each metric is \textbf{bold} and second best is \underline{underlined}.}
\label{table:prior_ablate}
\vskip 0.1in
\begin{center}
\begin{sc}
\begin{tabular}{c|cccccc}
\toprule
\multirow{2}{*}{\raisebox{-0.7\height}{\shortstack{Prior Distribution}}} 
& \multicolumn{3}{c}{Top-1} & \multicolumn{3}{c}{Top-10} \\
\cmidrule(lr){2-4}
\cmidrule(lr){5-7}
& Accuracy $\uparrow$ & MCES $\downarrow$ & Tanimoto $\uparrow$ & Accuracy $\uparrow$ & MCES $\downarrow$ & Tanimoto $\uparrow$ \\
\midrule
Fully Connected & 3.36\% & 12.67 & 0.28 & 7.60\% & 9.56 & 0.4 \\
Empty & \underline{6.60\%} & \textbf{11.55} & \underline{0.34} & \underline{14.94\%} & \textbf{9.07} & \textbf{0.47}\\
Marginal & \textbf{8.34\%} & \underline{11.95} & \textbf{0.35} & \textbf{15.44\%} & \underline{9.23} & \textbf{0.47} \\
\bottomrule
\end{tabular}
\end{sc}
\end{center}
\vskip -0.1in
\end{table*}

\section{Formulae Annotation Study}
\label{appenfix:formula_annotation}

In this section, we provide additional experiments using MIST-CF~\citep{goldman2023mist} and BUDDY~\citep{Xing2023-buddy} for formula annotation on the NPLIB1 and MassSpecGym datasets. Specifically, we use BUDDY and MIST-CF to predict the top-5 most likely formulae for each spectra in the test sets and measure the accuracy of these formula annotations. 

As shown in Table~\ref{table:buddy_mistcf_acc}, MIST-CF and BUDDY both achieve good performance on NPLIB1, where MIST-CF achieves over 90\% top-5 accuracy. However, both methods struggle on MassSpecGym, underscoring the difficulty of this dataset. It is important to note that neither NPLIB1 nor MassSpecGym include MS1 data, such as precursor m/z, which can aid in deriving accurate formula annotations. As such, these formula annotation accuracies are likely lower than what could be achieved in end-to-end elucidation workflows.

\begin{table*}[h]
\vspace{-0.1in}
\caption{Formula annotation accuracy for MIST-CF~\citep{goldman2023mist-cf} and BUDDY\cite{Xing2023-buddy} on the NPLIB1~\citep{bocker2016fragmentation} and MassSpecGym~\citep{bushuiev2024massspecgymbenchmarkdiscoveryidentification} datasets. The best performing method for each metric is \textbf{bold}.}
\label{table:buddy_mistcf_acc}
\vskip 0.1in
\begin{center}
\begin{sc}
\begin{tabular}{c|cccc}
\toprule
\multirow{2}{*}{\raisebox{-0.7\height}{\shortstack{Model}}} 
& \multicolumn{2}{c}{NPLIB1} & \multicolumn{2}{c}{MassSpecGym} \\
\cmidrule(lr){2-3}
\cmidrule(lr){4-5}
& Top-1 Acc. & Top-5 Acc. & Top-1 Acc. & Top-5 Acc. \\
\midrule
BUDDY & 78\% & 83\% & \textbf{59\%} & \textbf{71\%}\\
MIST-CF & \textbf{84\%} & \textbf{92\%} & 48\% & 69\%\\
\bottomrule
\end{tabular}
\end{sc}
\end{center}
\vskip -0.1in
\end{table*}

\clearpage

\section{\ours Generated Molecules}
\label{appendix:molecules}

\subsection{NPLIB1 Molecules}

\begin{figure}[h]
\centering
\vspace{-0.1in}
\subfigure{\includegraphics[width=1.0\textwidth]{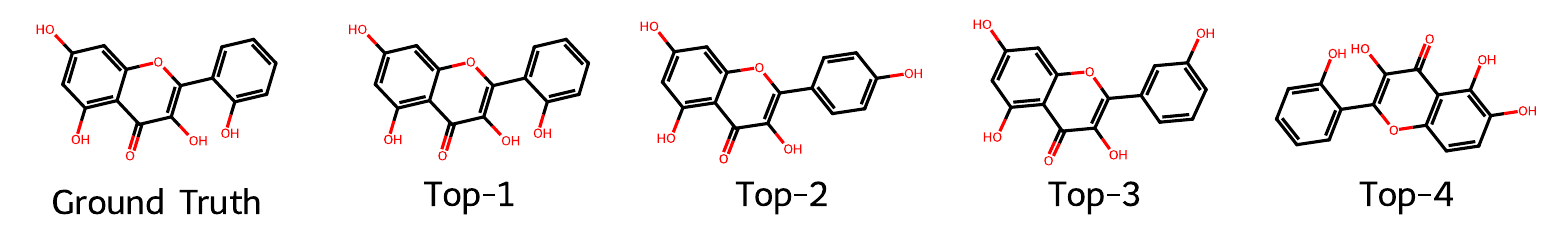}}
\subfigure{\includegraphics[width=1.0\textwidth]{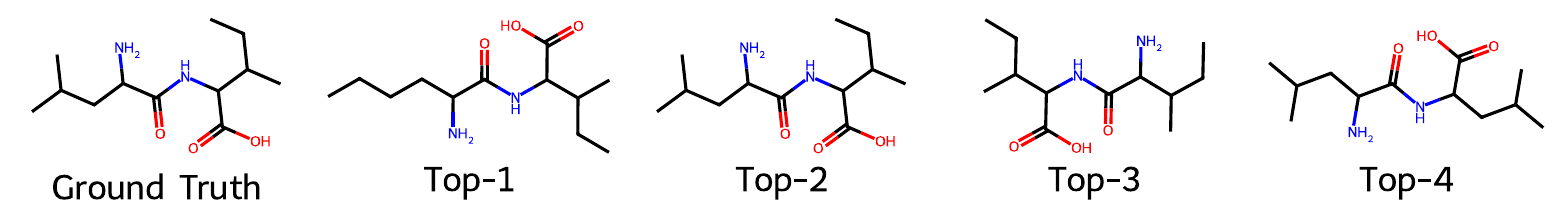}}
\subfigure{\includegraphics[width=1.0\textwidth]{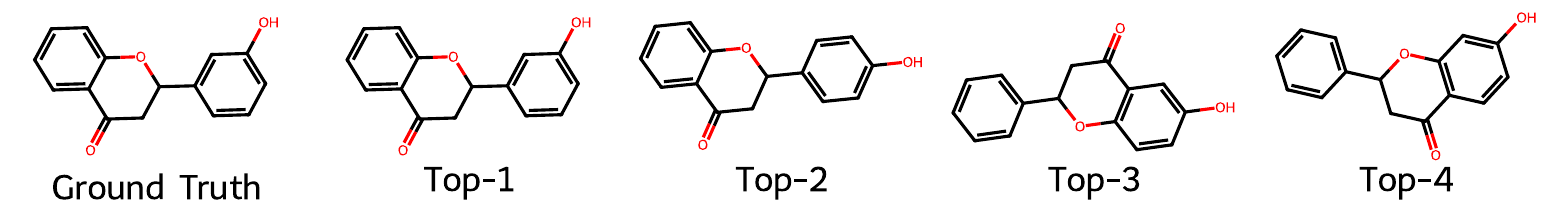}}
\subfigure{\includegraphics[width=1.0\textwidth]{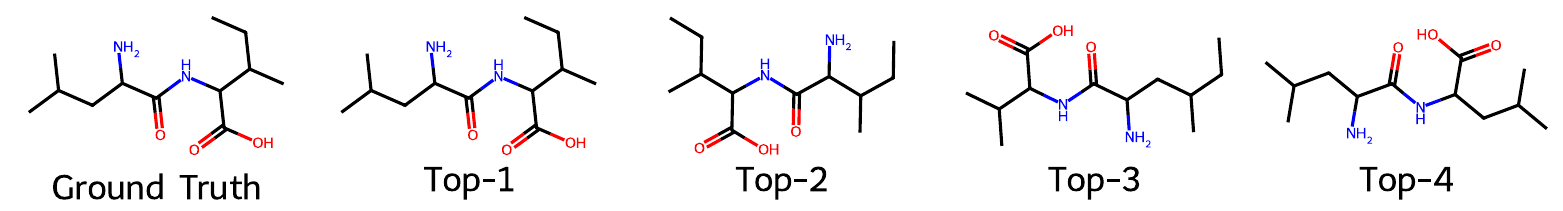}}
\subfigure{\includegraphics[width=1.0\textwidth]{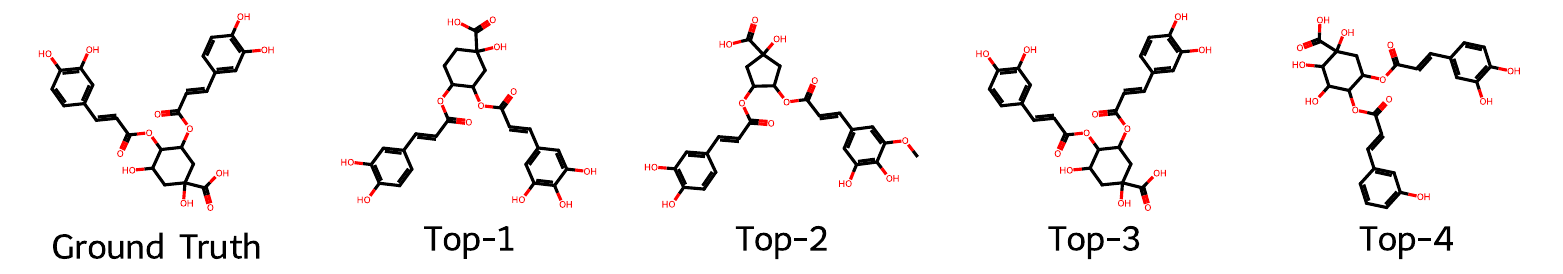}}
\subfigure{\includegraphics[width=1.0\textwidth]{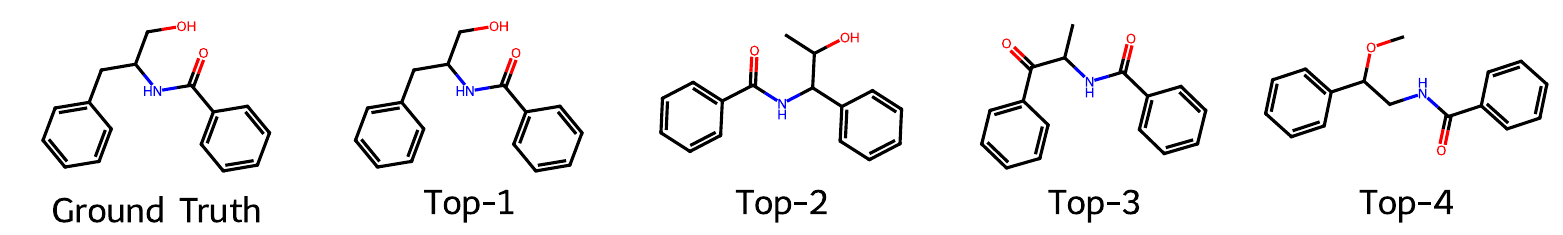}}
\subfigure{\includegraphics[width=1.0\textwidth]{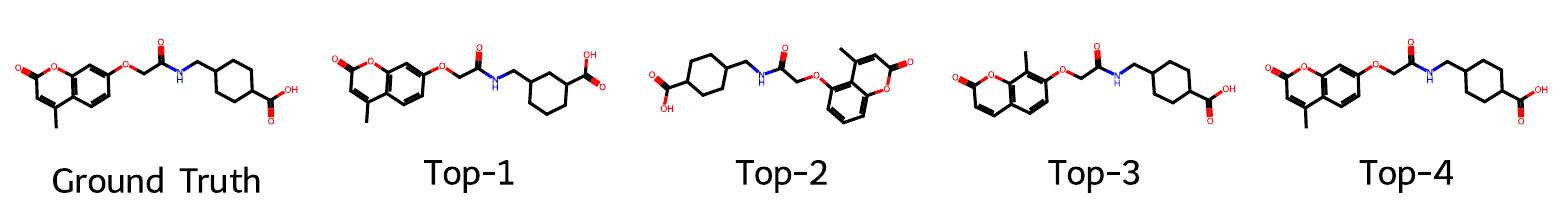}}

\vspace{-0.1in}
\caption{Positive (correct) test samples from the NPLIB1 dataset~\citep{duhrkop2021canopus}. Ground truth molecules (left column) and \ours predictions (right columns).}
\end{figure}

\begin{figure}[H]
\centering

\subfigure{\includegraphics[width=1.0\textwidth]{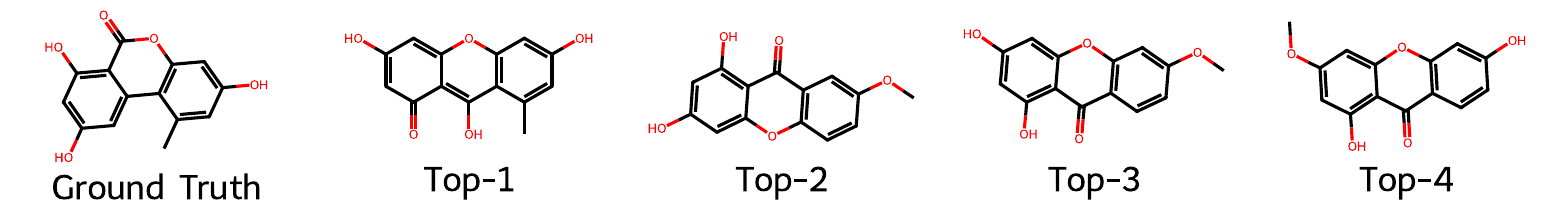}}
\subfigure{\includegraphics[width=1.0\textwidth]{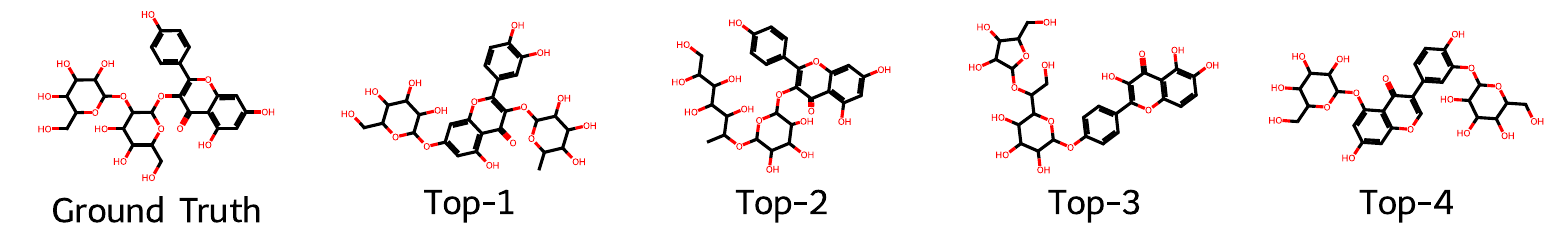}}
\subfigure{\includegraphics[width=1.0\textwidth]{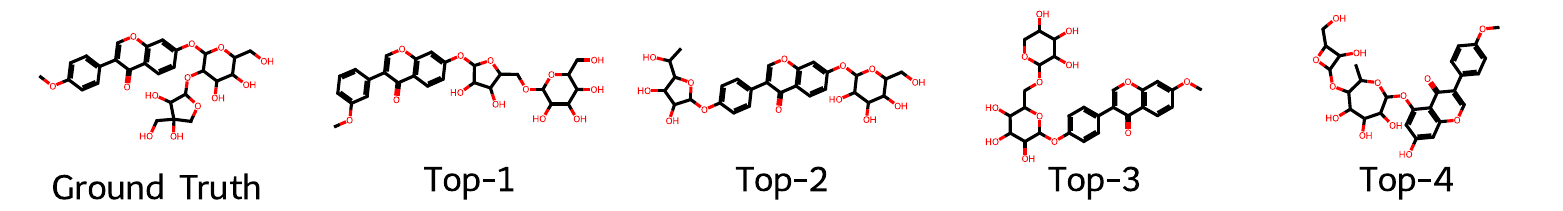}}
\subfigure{\includegraphics[width=1.0\textwidth]{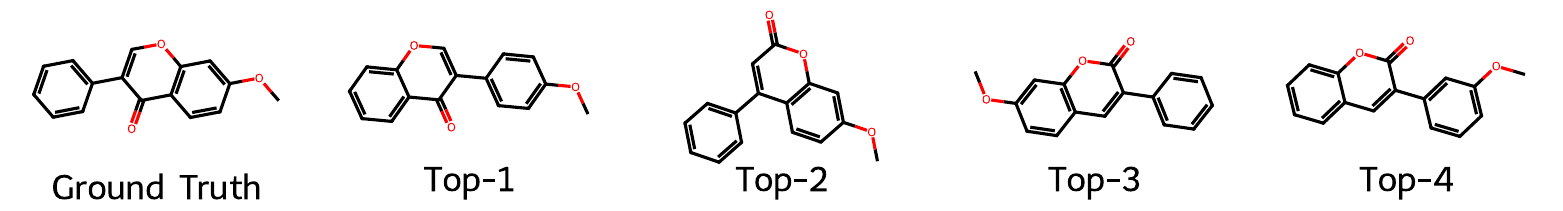}}
\subfigure{\includegraphics[width=1.0\textwidth]{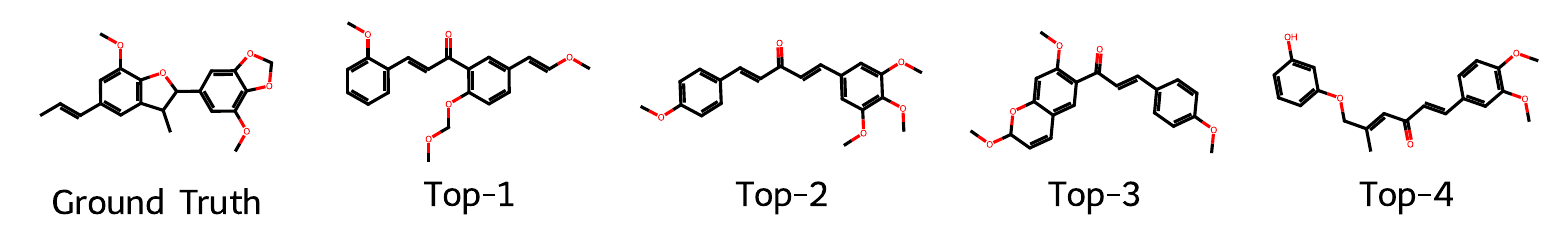}}
\subfigure{\includegraphics[width=1.0\textwidth]{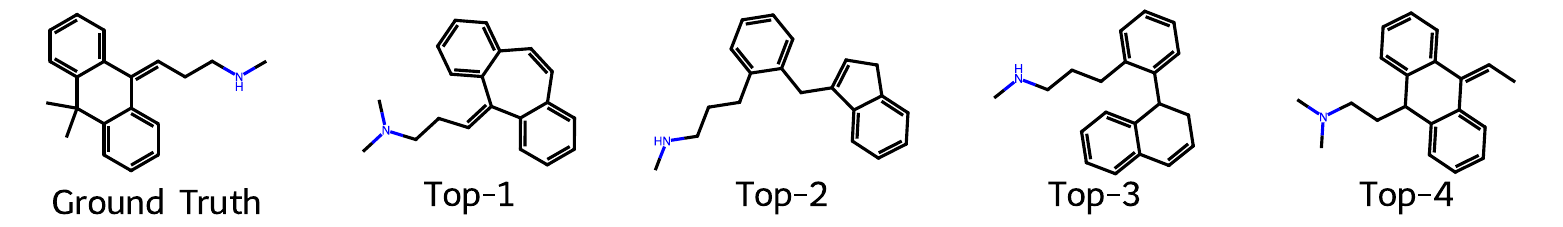}}
\subfigure{\includegraphics[width=1.0\textwidth]{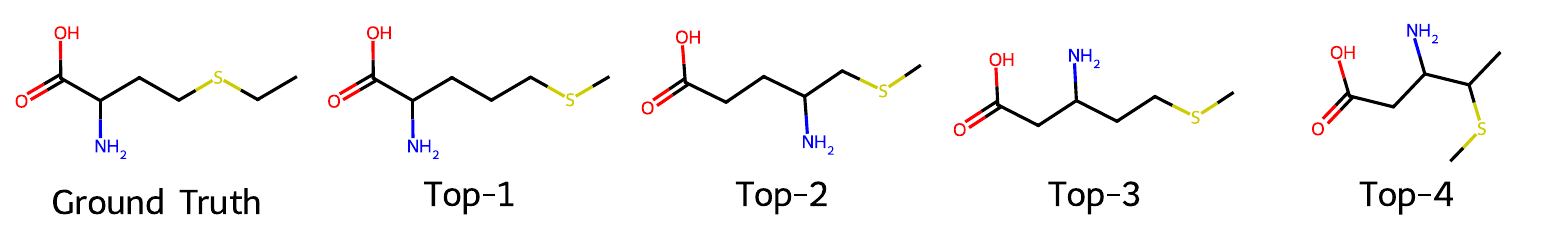}}

\vspace{-0.1in}
\caption{Negative (failure) test samples from the NPLIB1 dataset~\citep{duhrkop2021canopus}. Ground truth molecules (left column) and \ours predictions (right columns).}
\end{figure}

\clearpage

\subsection{MassSpecGym Molecules}

\begin{figure}[H]
\centering

\subfigure{\includegraphics[width=1.0\textwidth]{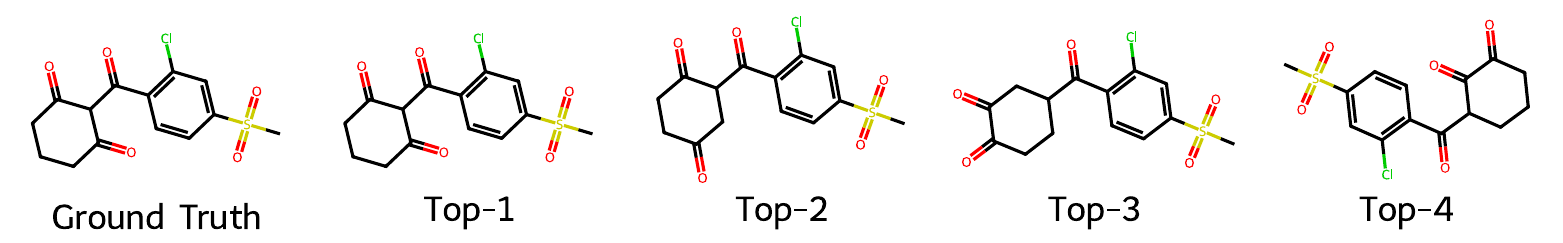}}
\subfigure{\includegraphics[width=1.0\textwidth]{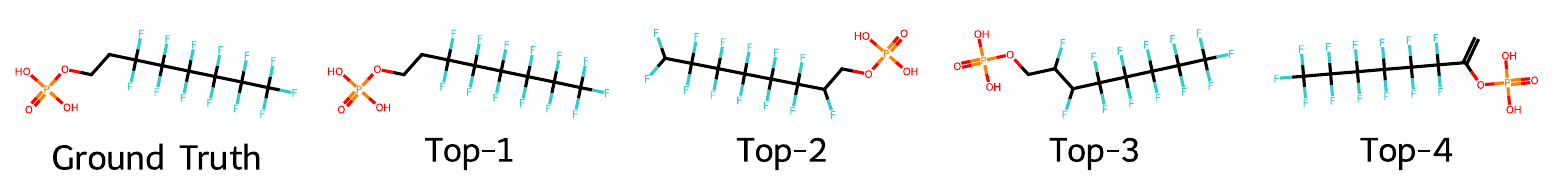}}
\subfigure{\includegraphics[width=1.0\textwidth]{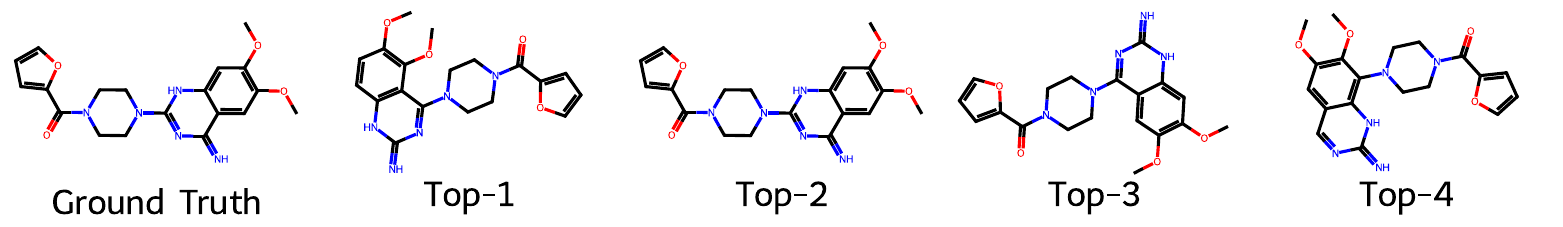}}
\subfigure{\includegraphics[width=1.0\textwidth]{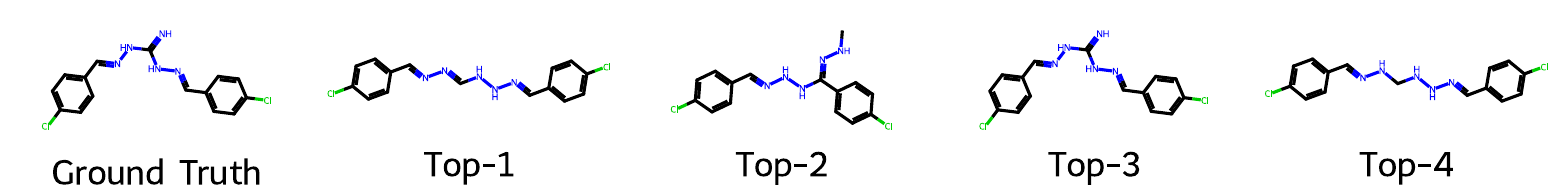}}
\subfigure{\includegraphics[width=1.0\textwidth]{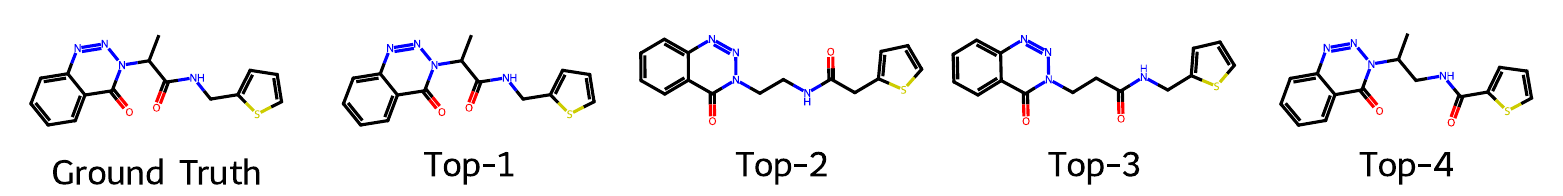}}
\subfigure{\includegraphics[width=1.0\textwidth]{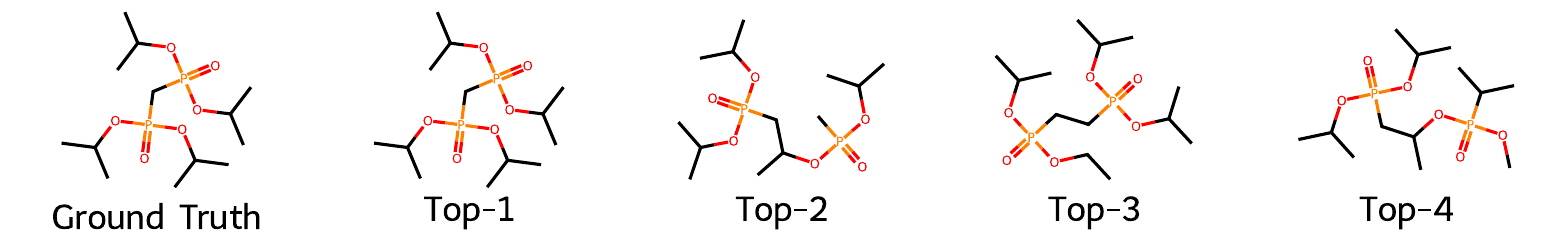}}
\subfigure{\includegraphics[width=1.0\textwidth]{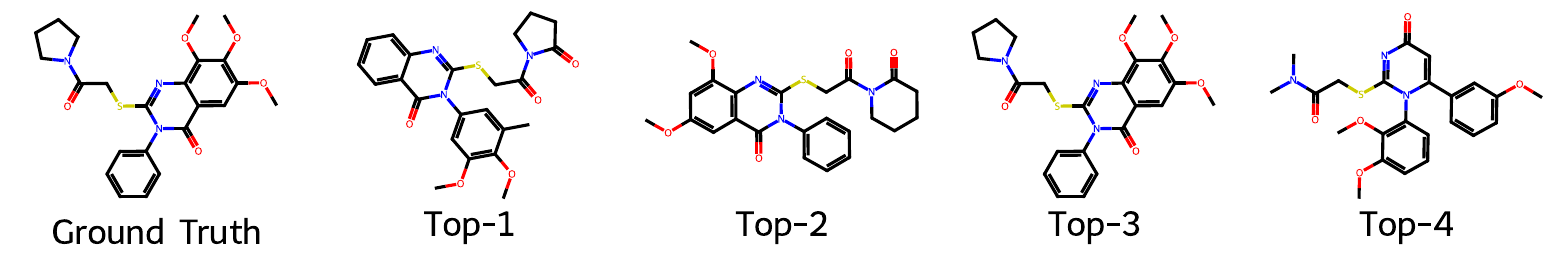}}

\vspace{-0.1in}
\caption{Positive (correct) test samples from the MassSpecGym dataset~\citep{bushuiev2024massspecgymbenchmarkdiscoveryidentification}. Ground truth molecules (left column) and \ours predictions (right columns).}
\end{figure}

\begin{figure}[h]
\centering

\subfigure{\includegraphics[width=1.0\textwidth]{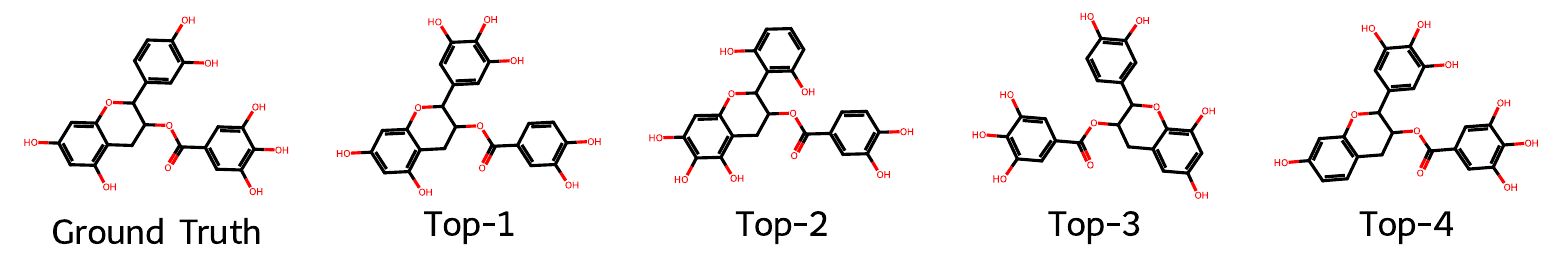}}
\subfigure{\includegraphics[width=1.0\textwidth]{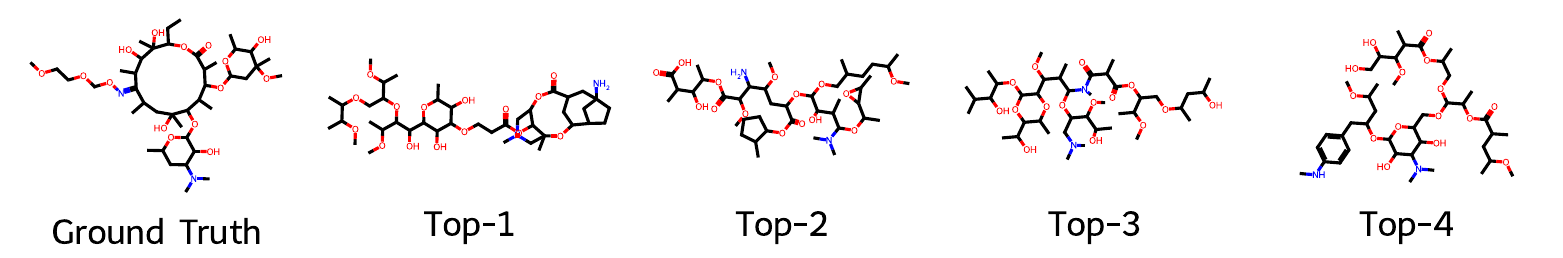}}
\subfigure{\includegraphics[width=1.0\textwidth]{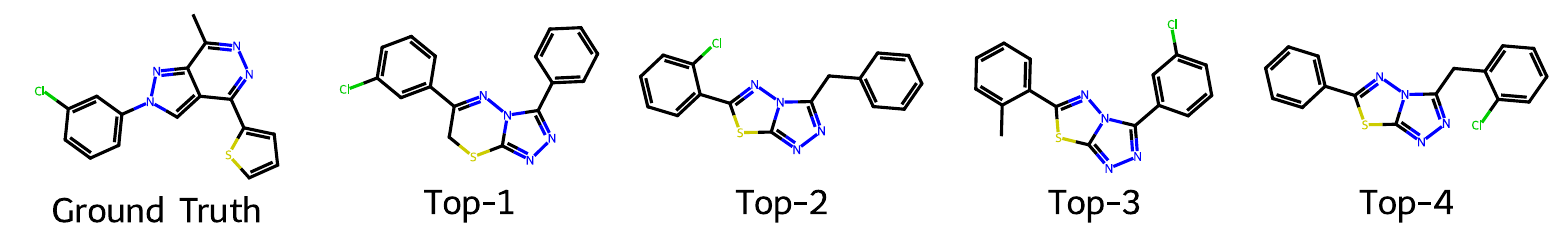}}
\subfigure{\includegraphics[width=1.0\textwidth]{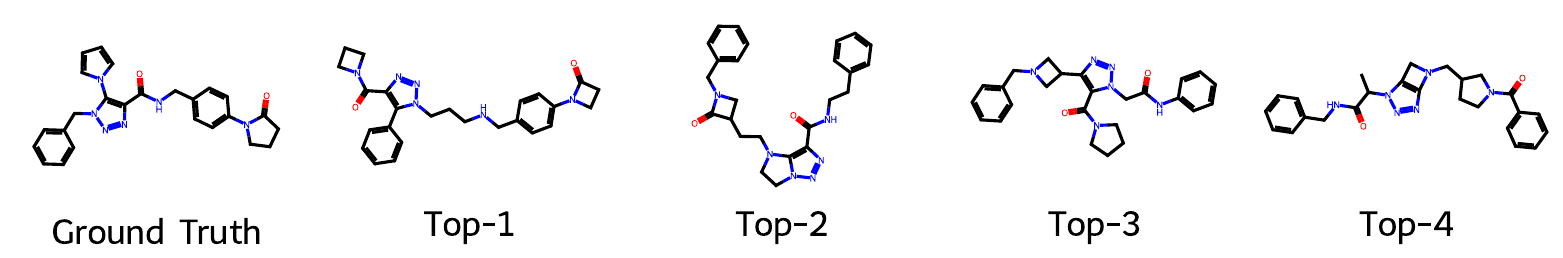}}
\subfigure{\includegraphics[width=1.0\textwidth]{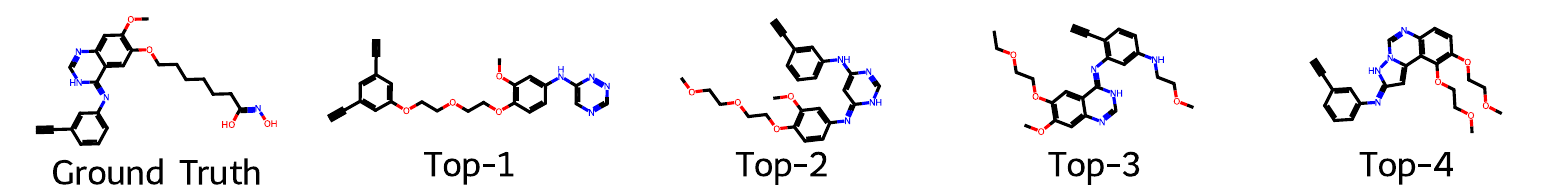}}
\subfigure{\includegraphics[width=1.0\textwidth]{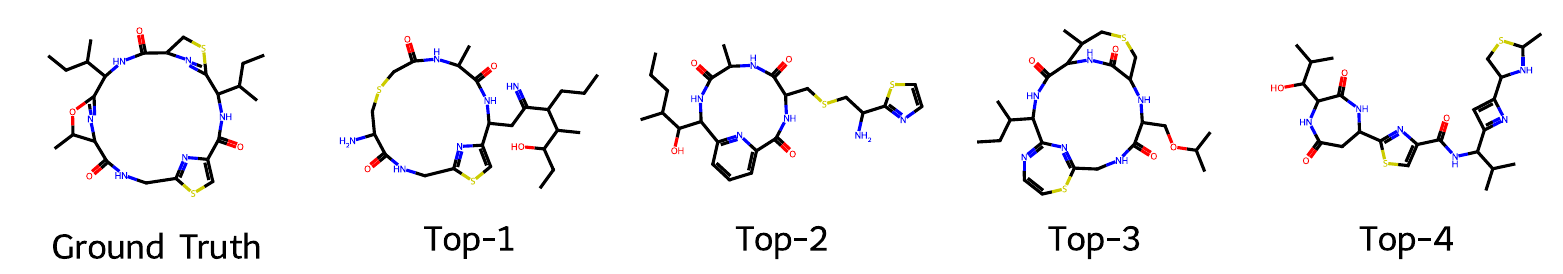}}
\subfigure{\includegraphics[width=1.0\textwidth]{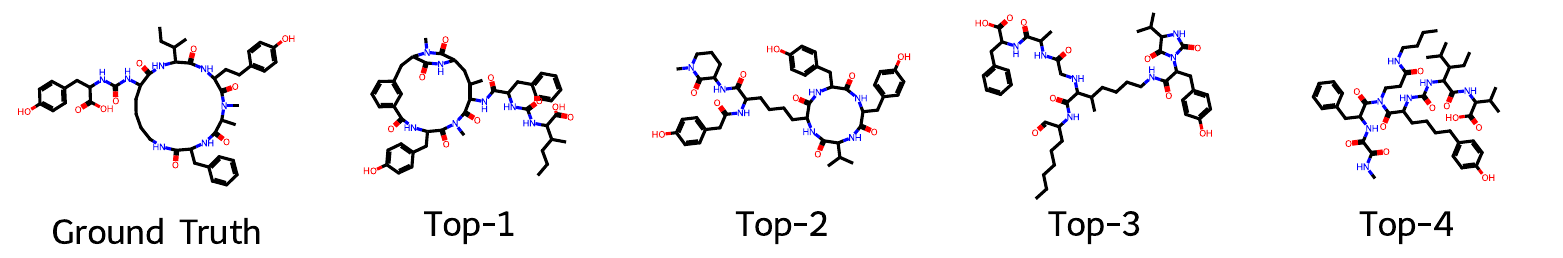}}

\vspace{-0.1in}
\caption{Negative (failure) test samples from the MassSpecGym dataset~\citep{bushuiev2024massspecgymbenchmarkdiscoveryidentification}. Ground truth molecules (left column) and \ours predictions (right columns).}
\end{figure}

\end{document}